# Fully Decentralized Multi-Agent Reinforcement Learning with Networked Agents

Kaiqing Zhang[♮]  Zhuoran Yang[†]  Han Liu[∗‡]  Tong Zhang[‡]  Tamer Başar[♮]

**Abstract**

We consider the problem of *fully decentralized* multi-agent reinforcement learning (MARL), where the agents are located at the nodes of a time-varying communication network. Specifically, we assume that the reward functions of the agents might correspond to different tasks, and are only known to the corresponding agent. Moreover, each agent makes individual decisions based on both the information observed locally and the messages received from its neighbors over the network.

Within this setting, the collective goal of the agents is to maximize the globally averaged return over the network through exchanging information with their neighbors. To this end, we propose two decentralized actor-critic algorithms with function approximation, which are applicable to large-scale MARL problems where both the number of states and the number of agents are massively large. Under the decentralized structure, the actor step is performed individually by each agent with no need to infer the policies of others. For the critic step, we propose a consensus update via communication over the network. Our algorithms are fully incremental and can be implemented in an online fashion. Convergence analyses of the algorithms are provided when the value functions are approximated within the class of linear functions. Extensive simulation results with both linear and nonlinear function approximations are presented to validate the proposed algorithms. Our work appears to be the first study of fully decentralized MARL algorithms for networked agents with function approximation, with provable convergence guarantees.

## 1   Introduction

In reinforcement learning (Sutton and Barto, 1998), the decision maker (synonymously, agent) aims to achieve the optimal behavior in the presence of uncertainty by interacting with the environment, which is usually modeled as a Markov Decision Process (MDP). With the advancement of deep learning (Goodfellow et al., 2016), reinforcement learning has been shown to achieve striking

---

[♮]Department of Electrical and Computer Engineering & Coordinated Science Laboratory, University of Illinois at Urbana-Champaign

[†]Department of Operations Research and Financial Engineering, Princeton University

[∗]Department of Electrical Engineering and Computer Science and Statistics, Northwestern University

[‡]Tencent AI Lab

performances in MDP problems such as board games (Silver et al., 2016, 2017), autonomous driving (Shalev-Shwartz et al., 2016), and robotics (Kober and Peters, 2012). See Li (2017) for an overview of recent achievements of deep reinforcement learning.

In this work, we study the problem of multi-agent reinforcement learning (MARL), where a common environment is influenced by the joint actions of multiple agents. In particular, at each state, each agent takes an action, and these actions together determine the next state of the environment and the reward of each agent. In addition, the agents are allowed to have different reward functions from different tasks, but each agent can observe only its own reward. We are interested in the collaborative setting where the agents have a common goal, which is to jointly maximize the globally averaged return over all agents in the environment.

For collaborative MARL problems, it is pivotal to specify the protocol of collaboration among the agents. One tempting choice is to have a central controller which receives the rewards of all agents, and determines the action for each agent. With information of all the agents available to the controller, the problem reduces to a classical MDP and can be solved by existing single-agent reinforcement learning algorithms. However, in many real-world scenarios, such as sensor networks (Rabbat and Nowak, 2004) and intelligent transportation systems (Adler and Blue, 2002), a central controller simply does not exist or may be costly to install. Moreover, the central controller needs to communicate with each agent to exchange the information, which incessantly increases the communication overhead at the single controller. This may degrade the *scalability* of the multi-agent system as well as its *robustness* to malicious attacks.

In view of all these disadvantages of the centralized protocol, we consider here a decentralized protocol where the agents are connected by a possibly time-varying communication network, which serves as the channels for the agents to exchange information in absence of any central controller. Specifically, let $\{\mathcal{G}_t = (\mathcal{N}, \mathcal{E}_t)\}_{t \geq 0}$ be a time-varying network, where $\mathcal{N}$ is the set of all nodes, and $\mathcal{E}_t \subseteq \{(i,j) \colon i,j \in \mathcal{N}\}$ is the set of all edges at time $t$. We assume that each node represents one agent and that two agents $i \in \mathcal{N}$ and $j \in \mathcal{N}$ can communicate with each other at time $t$ if and only if $(i,j) \in \mathcal{E}_t$. As such, at each time step, each agent executes an individual action based on both the local information and the message sent from its neighbors, with the joint goal of maximizing the average rewards of all agents over the network. We refer to this protocol as *networked multi-agent reinforcement learning*, which is presented in §2.2 in precise terms. This framework of networked architecture finds a broad range of applications in practical multi-agent systems, such as unmanned vehicles (Fax and Murray, 2004), robotics (Corke et al., 2005), power grid (Callaway and Hiskens, 2011), and mobile sensor networks (Cortes et al., 2004).

With only local reward and action, classical reinforcement learning algorithms can hardly maximize the networked-wide averaged reward determined by the joint actions of all agents. To resolve this problem, we propose two decentralized actor-critic algorithms for networked multi-agent systems, based on a novel policy gradient theorem for MARL. Specifically, the actor step is performed individually by each agent without the need to infer the policies of others. For the critic step, on the other hand, each agent shares its estimate of the value function with its neighbors on the network, so that a consensual estimate is achieved, which is further used in the subsequent actor step. In this regard, the local information at each agent is able to diffuse across the network, making the network-wide maximum reward achievable. Our algorithms enjoy the advantages of scalability to large population of agents, robustness against malicious attacks, and communication efficiency, as in standard distributed/decentralized algorithms over networked systems (Nedic and Ozdaglar, 2009; Ram et al., 2010; Agarwal and Duchi, 2011; Jakovetic et al., 2011; Tu and Sayed,

2012).

Furthermore, one long-standing problem in RL is to increase the scalability of algorithms with high dimensional state-action spaces. The problem becomes more pronounced in MARL since the number of joint actions grows exponentially with the number of agents in the system. To handle this exponential growth, we approximate both the policy and value functions by some parametrized function classes, such as deep neural networks. Thus, combining the decentralized network architecture and function approximation, our algorithms can be readily applied to large-scale MARL problems where both the number of states and the number of agents are massive. More importantly, we show that our algorithms converge when linear function approximation is used, which provides theoretical support for the proposed MARL framework with networked agents.

**Main Contribution.** Our contribution in this work is three-fold. First, we formulate the fully decentralized MARL problem for networked agents, and prove a version of the policy gradient theorem adapted to this setting. Second, we propose two decentralized actor-critic algorithms with function approximation, which enable our approach to be applied to large-scale MARL problems. Third, in the special case of where linear function approximation is used, we establish convergence guarantees for the proposed algorithms. It appears that this is the first class of fully decentralized MARL algorithms for networked agents with function approximation, with provable convergence guarantees.

**Related Work.** Our algorithms belong to the class of actor-critic algorithms. There is a huge body of literature on single-agent actor-critic algorithms, which are based on the policy gradient theorem (Baxter and Bartlett, 2000). The first actor-critic algorithm has been introduced in Sutton et al. (2000); Konda and Tsitsiklis (2000), which have also studied the convergence of the algorithm with linear function approximation. Later, Kakade (2002) has proposed to update the policy function using natural gradient descent (Amari, 1998), which leads to the natural actor-critic algorithm (Peters and Schaal, 2008). Convergence of actor-critic and natural actor-critic algorithms with linear function approximation have been studied in Bhatnagar et al. (2009, 2008); Castro and Meir (2010); Bhatnagar (2010). Recently, for deep reinforcement learning, where deep neural networks are used for function approximation, various actor-critic algorithms have been proposed. For example, Gruslys et al. (2017); Wang et al. (2016); Gu et al. (2017) have proposed sample-efficient actor-critic algorithms based on experience replay (Mnih et al., 2015) and off-policy learning (Munos et al., 2016). In addition, for continuous action spaces, Silver et al. (2014); Lillicrap et al. (2016) have proposed actor-critic algorithms based on deterministic policies. A more related and popular work is Mnih et al. (2016), which has proposed the asynchronous actor-critic (A3C) algorithm. Unlike our MARL framework, the A3C algorithm essentially deals with single-agent RL but with multiple parallel workers/processors, where no control interaction occurs among the workers. Moreover, a central controller is required to coordinate the asynchronous update of the workers. In contrast, our actor-critic algorithms are built upon a version of policy gradient theorem for MARL, with no central controller required in implementation.

A line of research more relevant to our work here is on MARL. Most existing work on MARL is based on the framework of Markov games, which was first proposed by Littman (1994) and then followed by Littman (2001); Lauer and Riedmiller (2000); Hu and Wellman (2003). This framework applies to the setting with both collaborative and competitive relationships among agents. However, these early algorithms have been developed only for tabular cases where no function approximation is used. Moreover, for the collaborative setting considered here, notably team games, references Wang and Sandholm (2003); Arslan and Yüksel (2017) take the reward of all agents to be identical, which greatly simplifies the problem since the value function can be estimated locally with no need of information exchange among agents. More recently, several MARL algorithms using deep neural networks as function approximators have gained increasing attention (Foerster et al., 2016; Gupta et al., 2017; Lowe et al., 2017; Omidshafiei et al., 2017; Foerster et al., 2017). However, most of them focus on showing empirical results, without much convergence guarantees. Moreover, none of them have been designed for our MARL framework with networked agents, where the communication among agents may contribute toward the overall performance of MARL in a fully decentralized setting. See §C for a detailed comparison with existing MARL models and algorithms.

Since we assume the reward functions of the agents to be possibly different, which might correspond to various different tasks, our work is also pertinent to the literature on multi-task reinforcement learning (Wilson et al., 2007; Parisotto et al., 2015; Teh et al., 2017), which focuses on scenarios where a single-agent solves multiple related MDP problems. Common knowledge is distilled and transferred among the training of policies for different tasks. Although multiple workers can be used to learn different policies by sharing knowledge with each other (Teh et al., 2017), the workers solve relevant but independent MDPs, where the workers do not interact with each other as opposed to our framework. In fact, we consider multiple agents to be heterogeneous with distinct policies and rewards, while some multi-task RL algorithms rely heavily on the assumption that the polices of various tasks are similar to a great extent.

**Notation.** For any vector $x \in \mathbb{R}^n$ and matrix $Y \in \mathbb{R}^{m \times n}$, we use $\|x\|$ and $\|Y\|$ to denote the Euclidean norm of $x$ and the induced 2-norm of $Y$, respectively. We also use $\|x\|_\infty$ and $\|Y\|_\infty$ to denote the infinite norm of $x$ and the induced infinite-norm of $Y$, respectively. For a finite set $\mathcal{A}$, we use $|\mathcal{A}|$ to denote its cardinality. For notational simplicity, we use $\lim_t$, $\sup_t$, and $\sum_t$ to represent $\lim_{t \to \infty}$, $\sup_{t \to \infty}$, and $\sum_{t \geq 0}$, respectively. For any two numbers $a$ and $b$, we write $a \ll b$ if $a$ is much smaller than $b$. If not otherwise specified, we denote by I and $\mathbb{1}$ the identity matrix and all-one vector of proper dimensions, respectively. We also use $[N]$ to denote the set of integers $\{1, \cdots, N\}$ and $\mathbb{N}$ to denote the set of non-negative integers.

**Roadmap.** In §2 we first provide the background material on reinforcement learning, and then formulate the networked multi-agent MDP to be studied in this work. We present two decentralized actor-critic algorithms for fully decentralized MARL problems in §3 and provide theoretical convergence guarantees in §4 when linear function approximation is applied. Proofs of the main results are given in §5. Experiments with linear function approximation are presented in §6 to back up our theory.

## 2 Background

In this section, we introduce the background and formulation of the networked multi-agent reinforcement learning problem.

### 2.1 Markov Decision Process and Actor-critic Algorithm

A Markov decision processes is characterized by a quadruple $\mathcal{M} = \langle \mathcal{S}, \mathcal{A}, P, r \rangle$, where $\mathcal{S}$ is a finite state space, $\mathcal{A}$ is a finite action space, $P(s'|s,a) : \mathcal{S} \times \mathcal{A} \times \mathcal{S} \to [0,1]$ is a state transition probability from state $s$ to $s'$ determined by action $a$, and $R(s,a) : \mathcal{S} \times \mathcal{A} \to \mathbb{R}$ is a reward function defined by $R(s,a) = \mathbb{E}[r_{t+1}|s_t = s, a_t = a]$, with $r_{t+1}$ being the instantaneous reward at time $t$. Policy of the agent is a mapping $\pi : \mathcal{S} \times \mathcal{A} \to [0,1]$, representing the probability of choosing action $a$ at state $s$. The objective of the agent is to find the optimal policy that maximizes the expected time-average reward, notably, long-term return, which is given by $J(\pi)$:

$$J(\pi) = \lim_T \frac{1}{T} \sum_{t=0}^{T-1} \mathbb{E}(r_{t+1}) = \sum_{s \in \mathcal{S}} d_\pi(s) \sum_{a \in \mathcal{A}} \pi(s,a) R(s,a), \tag{2.1}$$

where $d_\pi(s) = \lim_t \mathbb{P}(s_t = s|\pi)$ is the stationary distribution of the Markov chain under policy $\pi$. Note that the distribution $d_\pi(s)$ and the limit in (2.1) are well defined under the assumption that with any policy $\pi$, the Markov chain resulting from the MDP is irreducible and aperiodic.

Given any policy $\pi$, the relative action-value associated with a state-action pair $(s,a)$ is thus defined as (Sutton and Barto, 1998; Puterman, 2014)

$$Q_\pi(s,a) = \sum_t \mathbb{E}[r_{t+1} - J(\pi)|s_0 = s, a_0 = a, \pi].$$

Accordingly, the relative state-value associated with state $s$ under policy $\pi$ can be defined as $V_\pi(s) = \sum_{a \in \mathcal{A}} \pi(s,a) Q_\pi(s,a)$. For simplicity, we will hereafter refer to $V_\pi$ and $Q_\pi(s,a)$ as *state-value* function and *action-value* function, respectively. When the state and action space is massively large, $V_\pi$ and $Q_\pi(s,a)$ are usually approximated by some parametrized functions $Q(\cdot,\cdot;\omega)$ and $V(\cdot;v)$, respectively, with parameters $\omega$ and $v$. The policy $\pi$ can also be parametrized as $\pi_\theta$. For notational convenience, hereafter we denote all $\pi_\theta$ in the subscript of other symbols as $\theta$, e.g., we write $d_{\pi_\theta}$ and $V_{\pi_\theta}$ as $d_\theta$ and $V_\theta$, respectively.

Under this parameterization, actor-critic (AC) algorithms have been advocated to solve for the optimal policy $\pi_\theta$. In particular, the actor-critic algorithms are based on the well-known policy gradient theorem (Sutton et al., 2000), where the gradient of the return $J(\theta)$ with respect to the policy parameter $\theta$ can be written as

$$\nabla_\theta J(\theta) = \mathbb{E}_{s \sim d_\theta, a \sim \pi_\theta} \{\nabla_\theta \log \pi_\theta(s,a) \cdot [Q_\theta(s,a) - b(s)]\}.$$

The term $b(s)$ is usually referred to as the baseline, and $\nabla_\theta \log \pi_\theta$ is referred to as the *score function* of the policy $\pi_\theta$. It has been recognized in Bhatnagar et al. (2008) that the minimum variance baseline in the action-value function estimator corresponds to the state-value function $V_\theta(s)$. Let

$$A_\theta(s,a) = Q_\theta(s,a) - V_\theta(s) \tag{2.2}$$

be the *advantage* function. At time step $t$, define $Q_t(\omega) = Q(s_t, a_t; \omega)$ and let $A_t$ be the sample of the advantage function, i.e.,

$$A_t = Q(s_t, a_t; \omega_t) - \sum_{a \in \mathcal{A}} \pi_{\theta_t}(s_t, a) Q(s_t, a; \omega_t)$$

Let $\psi_t = \nabla_\theta \log \pi_{\theta_t}(s_t, a_t)$ represent the sample of the score function. As such, one common AC algorithm based on action-value function approximation has the following form

$$\mu_{t+1} = (1 - \beta_{\omega,t}) \cdot \mu_t + \beta_{\omega,t} \cdot r_{t+1}, \quad \omega_{t+1} = \omega_t + \beta_{\omega,t} \cdot \delta_t \cdot \nabla_\omega Q_t(\omega_t), \quad \theta_{t+1} = \theta_t + \beta_{\theta,t} \cdot A_t \cdot \psi_t, \tag{2.3}$$

where $\beta_{\omega,t}, \beta_{\theta,t} > 0$ are stepsizes, $\mu_t$ tracks the unbiased estimate of the average return, and

$$\delta_t = r_{t+1} - \mu_t + Q(s_{t+1}, a_{t+1}; \omega_t) - Q(s_t, a_t; \omega_t)$$

denotes the action-value temporal difference (TD) error. Note that the action $a_{t+1}$ is sampled from the policy $\pi_{\theta_t}(s_{t+1}, \cdot)$. The AC algorithm is usually developed as a two-time-scale algorithm, by setting the stepsizes $\beta_{\omega,t}$ and $\beta_{\theta,t}$ following $\lim_t \beta_{\theta,t} \cdot \beta_{\omega,t}^{-1} = 0$. Thus the first two updates in (2.3), referred to as the *critic step*, operate at a faster time scale to estimate the action-value function $Q(s_t, a_t; \omega_t)$ under policy $\pi_{\theta_t}$. The last update in (2.3), on the other hand, corresponds to the so-called *actor step*, which improves the policy along the gradient ascent direction at a slower time scale.

Furthermore, actor-critic algorithms are shown to achieve state-of-art performances in many complicated application domains (Peters and Schaal, 2008; Schulman et al., 2015; Mnih et al., 2016; Bahdanau et al., 2016; Silver et al., 2017). Driven by these success stories, we focus on designing actor-critic algorithms for MARL over networks in the present work.

## 2.2 Multi-Agent Reinforcement Learning

Consider now a system of $N$ agents operating in a common environment, denoted by $\mathcal{N} = [N]$. We are interested in the decentralized setting where no central controller exists in the system that either collects rewards or makes the decisions for the agents. Instead, the agents are located on a possibly time-varying communication network, denoted by $\mathcal{G}_t = (\mathcal{N}, \mathcal{E}_t)$, where $\mathcal{E}_t$ represents the set of communication links at time $t \in \mathbb{N}$. In other words, $\mathcal{G}_t$ is a time-varying and undirected graph with vertex set $\mathcal{N}$ and edge set $\mathcal{E}_t \subseteq \{(i,j): i, j \in \mathcal{N}, i \neq j\}$ at time $t$, where an edge $(i,j) \in \mathcal{E}_t$ means that agents $i$ and $j$ can share information at time $t$[1]. We define the multi-agent MDP over communication networks in detail as follows, which is a natural extension of classical MDP to the decentralized multi-agent network.

**Definition 2.1** (Networked Multi-Agent MDP). Let $\{\mathcal{G}_t = (\mathcal{N}, \mathcal{E}_t)\}_{t \geq 0}$ be a time-varying communication network with $N$ agents. A networked multi-agent MDP is characterized by a tuple $(\mathcal{S}, \{\mathcal{A}^i\}_{i \in \mathcal{N}}, P, \{R^i\}_{i \in \mathcal{N}}, \{\mathcal{G}_t\}_{t \geq 0})$ where $\mathcal{S}$ is the global state space shared by all the agents in $\mathcal{N}$, and $\mathcal{A}^i$ is the set of actions that agent $i$ can execute. In addition, let $\mathcal{A} = \prod_{i=1}^N \mathcal{A}^i$ be the joint action space of all agents. Then, $R^i: \mathcal{S} \times \mathcal{A} \to \mathbb{R}$ is the local reward function of agent $i$, and $P: \mathcal{S} \times \mathcal{A} \times \mathcal{S} \to [0,1]$ is the state transition probability of the MDP. Moreover, we assume that the states and the joint actions are globally observable whereas the rewards are observed only locally.

In view of this definition, at time step $t$, suppose that the global state is $s_t \in \mathcal{S}$ and the agents execute joint actions $a_t = (a_t^1, \ldots, a_t^N) \in \mathcal{A}$. As a result, each agent $i \in \mathcal{N}$ receives a reward $r_{t+1}^i$, which is a random variable with expected value $R^i(s_t, a_t)$. Moreover, the MDP shifts to a new state $s_{t+1} \in \mathcal{S}$ with probability $P(s_{t+1} | s_t, a_t)$. We say our model is *fully decentralized* since both the reward is received locally and the action is executed locally by each agent.

Furthermore, since each agent chooses its own action individually at each time, it is reasonable to assume that their choices of actions are conditionally independent given the current state. More specifically, let $\pi: \mathcal{S} \times \mathcal{A} \to [0,1]$ be a joint policy function, that is, $\pi(s,a)$ is the probability of choosing action $a$ at state $s$. We assume that $\pi(s,a) = \prod_{i \in \mathcal{N}} \pi^i(s, a^i)$, where $\pi^i(s, a^i)$ is the probability that agent $i$ selects action $a^i \in \mathcal{A}^i$ at state $s$. In other words, the joint policy $\pi$ can be factorized as the product of local policies $\{\pi^i : \mathcal{S} \times \mathcal{A}^i \to [0,1]\}_{i \in \mathcal{N}}$, i.e., the randomized actions picked by different agents are statistically independent.

---
[1] Here we treat $(i,j)$ and $(j,i)$ as the same, which represent the edge between nodes $i$ and $j$.

Moreover, when the state space $\mathcal{S}$ is large, it is useful to consider policies that are in a parametric function class. For any agent $i$, we assume that the local policy is given by $\pi^i_{\theta^i}$, where $\theta^i \in \Theta^i$ is the parameter, and $\Theta^i \subseteq \mathbb{R}^{m_i}$ is a compact set. We pack the parameters together by writing $\theta = [(\theta^1)^\top, \cdots, (\theta^N)^\top]^\top \in \Theta$, where $\Theta = \prod_{i=1}^N \Theta^i$. Then the joint policy is given by $\pi_\theta(s, a) = \prod_{i \in \mathcal{N}} \pi^i_{\theta^i}(s, a_i)$. In the following, we make an regularity assumption on the networked MDP and policy function, which is standard in the literature.

**Assumption 2.2.** We assume that for any $i \in \mathcal{N}$, $s \in \mathcal{S}$, and $a^i \in \mathcal{A}^i$, the policy function $\pi^i_{\theta^i}(s, a^i) > 0$ for any $\theta^i \in \Theta^i$. Also, $\pi^i_{\theta^i}(s, a^i)$ is continuously differentiable with respect to the parameter $\theta^i$ over $\Theta^i$. In addition, for any $\theta \in \Theta$, let $P^\theta$ be the transition matrix of the Markov chain $\{s_t\}_{t \geq 0}$ induced by policy $\pi_\theta$, that is,

$$P^\theta(s'|s) = \sum_{a \in \mathcal{A}} \pi_\theta(s,a) \cdot P(s'|s,a), \quad \forall s, s' \in \mathcal{S}. \tag{2.4}$$

We assume that the Markov chain $\{s_t\}_{t \geq 0}$ is irreducible and aperiodic under any $\pi_\theta$.

Assumption 2.2 is standard in the early work on AC algorithms with function approximation (Konda and Tsitsiklis, 2000; Bhatnagar et al., 2009). It is assumed that the policy $\pi_\theta(\cdot,\cdot)$ is continuously differentiable with respect to $\theta$, which is required by policy gradient methods and is satisfied by well-known function classes such as deep neural networks. Moreover, since the Markov chain induced by $\pi_\theta$ is irreducible and aperiodic, it has a stationary distribution of $d_\theta(s)$ over $\mathcal{S}$. Likewise, the Markov chain of the state-action pair $\{(s_t, a_t)\}_{t \geq 0}$ has a stationary distribution $d_\theta(s) \cdot \pi_\theta(s, a)$ for any $s \in \mathcal{S}$ and $a \in \mathcal{A}$.

In addition, the joint objective of the agents is to collaboratively find a joint policy $\pi_\theta$ that maximizes the *globally* averaged long-term return over the network based solely on *local* information. Recall that at time $t$, agent $i$ receives reward $r^i_{t+1}$ and the reward function of agent $i$ is given by $R^i(\cdot,\cdot)$. Then our goal is to solve the optimization problem

$$\underset{\theta}{\text{maximize}} \quad J(\theta) = \lim_T \frac{1}{T} \mathbb{E}\left(\sum_{t=0}^{T-1} \frac{1}{N} \sum_{i \in \mathcal{N}} r^i_{t+1}\right) = \sum_{s \in \mathcal{S}} d_\theta(s) \sum_{a \in \mathcal{A}} \pi_\theta(s,a) \cdot \overline{R}(s,a), \tag{2.5}$$

where $\overline{R}(s,a) = N^{-1} \cdot \sum R^i(s,a)$ is the globally averaged reward function. Let $\overline{r}_t = N^{-1} \cdot \sum_{i \in \mathcal{N}} r^i_t$; then, we have $\overline{R}(s,a) = \mathbb{E}[\overline{r}_{t+1} | s_t = s, a_t = a]$. Accordingly, the global expected action value function associated with a state-action pair $(s,a)$ under policy $\pi_\theta$ becomes

$$Q_\theta(s,a) = \sum_t \mathbb{E}\big[\overline{r}_{t+1} - J(\theta) | s_0 = s, a_0 = a, \pi_\theta\big]. \tag{2.6}$$

Moreover, the global state-value function $V_\theta(s)$ is defined as $V_\theta(s) = \sum_{a \in \mathcal{A}} \pi_\theta(s,a) Q_\theta(s,a)$ accordingly.

This framework of networked multi-agent systems finds a broad range of applications in distributed cooperative control problems, including formation control of unmanned vehicles (Fax and Murray, 2004), cooperative navigation of robots (Corke et al., 2005), load management in energy networks (Callaway and Hiskens, 2011), and flocking of mobile sensor networks (Cortes et al., 2004), etc. However, most of the existing work on networked cooperative multi-agent systems is approached in a static setting, in the sense that the optimization objective is deterministic and there is no control input affecting the transition of the system, see efforts in Nedic and Ozdaglar (2009); Ram et al. (2010); Agarwal and Duchi (2011); Jakovetic et al. (2011); Tu and Sayed (2012). Our framework extends the above to a dynamic setting by explicitly modeling an MDP involving networked agents with only local sensing and control capabilities.

## 3 Multi-Agent Actor-Critic with Networked Agents

In this section, we present the proposed actor-critic algorithms for the multi-agent MDP with networked agents. We first establish a policy gradient theorem for MARL, which characterizes the gradient of $J(\theta)$ defined in (2.5) in closed form.

**Theorem 3.1** (Policy Gradient Theorem for MARL). *For any $\theta \in \Theta$, let $\pi_\theta \colon \mathcal{S} \times \mathcal{A} \to [0,1]$ be a policy and let $J(\theta)$ be the globally long-term averaged return defined in (2.5). In addition, let $Q_\theta$ and $A_\theta$ be the action-value function and advantage function defined in (2.6) and (2.2), respectively. Moreover, for any $i \in \mathcal{N}$, we define the local advantage function $A_\theta^i \colon \mathcal{S} \times \mathcal{A} \to \mathbb{R}$ as*

$$A_\theta^i(s,a) = Q_\theta(s,a) - \widetilde{V}_\theta^i(s,a^{-i}), \quad \text{where} \quad \widetilde{V}_\theta^i(s,a^{-i}) = \sum_{a^i \in \mathcal{A}^i} \pi_{\theta^i}^i(s,a^i) \cdot Q_\theta(s,a^i,a^{-i}). \tag{3.1}$$

*Here we denote by $a^{-i}$ the actions of all agents except for $i$. Recall that $\theta = [(\theta^1)^\top, \ldots, (\theta^N)^\top]^\top$. Then the gradient of $J(\theta)$ with respect to $\theta^i$ is given by*

$$\nabla_{\theta^i} J(\theta) = \mathbb{E}_{s \sim d_\theta, a \sim \pi_\theta}\left[\nabla_{\theta^i} \log \pi_{\theta^i}^i(s,a^i) \cdot Q_\theta(s,a)\right] = \mathbb{E}_{s \sim d_\theta, a \sim \pi_\theta}\left[\nabla_{\theta^i} \log \pi_{\theta^i}^i(s,a^i) \cdot A_\theta(s,a)\right]$$
$$= \mathbb{E}_{s \sim d_\theta, a \sim \pi_\theta}\left[\nabla_{\theta^i} \log \pi_{\theta^i}^i(s,a^i) \cdot A_\theta^i(s,a)\right]. \tag{3.2}$$

*Proof.* The proof of this theorem follows the proof of the policy gradient theorem in single-agent reinforcement learning (Sutton et al., 2000), which implies that

$$\nabla_\theta J(\theta) = \mathbb{E}_{s \sim d_\theta, a \sim \pi_\theta}[\nabla_\theta \log \pi_\theta(s,a) \cdot Q_\theta(s,a)]$$
$$= \sum_{s \in \mathcal{S}} d_\theta(s) \sum_{a \in \mathcal{A}} \pi_\theta(s,a) \left[\nabla_\theta \sum_{i \in \mathcal{N}} \log \pi_{\theta^i}^i(s,a^i)\right] \cdot Q_\theta(s,a), \tag{3.3}$$

where $Q_\theta$ is the action-value function defined in (2.6), and $d_\theta$ denotes the stationary distribution of the Markov chain induced by policy $\pi_\theta$. Here the second equality in (3.3) holds because $\pi_\theta$ is the product of local policy functions. Hence, the gradient with respect to each parameter $\theta^i$ becomes

$$\nabla_{\theta^i} J(\pi_\theta) = \sum_{s \in \mathcal{S}} d_\theta(s) \sum_{a \in \mathcal{A}} \pi_\theta(s,a) \cdot \nabla_{\theta^i} \log \pi_{\theta^i}^i(s,a^i) \cdot Q_\theta(s,a), \tag{3.4}$$

which proves the first equality in (3.2). Moreover, since $\sum_{a^i \in \mathcal{A}^i} \pi_{\theta^i}^i(s,a^i) = 1$, we have

$$\nabla_{\theta^i}\left[\sum_{a^i \in \mathcal{A}^i} \pi_{\theta^i}^i(s,a^i)\right] = 0. \tag{3.5}$$

To simplify the notation, for each $i \in \mathcal{N}$, we define $a^{-i}$ as the joint actions of all agents except $i$, and let $\mathcal{A}^{-i} = \prod_{j \neq i} \mathcal{A}^j$. Thus, for any function $F \colon \mathcal{S} \times \mathcal{A}^{-i} \to \mathbb{R}$ which does not rely on $a^i$, by (3.5), for any $s \in \mathcal{S}$ we have

$$\sum_{a \in \mathcal{A}} \pi_\theta(s, a) \cdot \bigl[\nabla_{\theta^i} \log \pi_{\theta^i}^i(s, a^i)\bigr] \cdot F(s, a^{-i})$$
$$= \sum_{a^{-i} \in \mathcal{A}^{-i}} F(s, a^{-i}) \cdot \Bigl[\prod_{j \in \mathcal{N}, j \neq i} \pi_{\theta^j}^j(s, a^j)\Bigr] \cdot \Bigl[\sum_{a^i \in \mathcal{A}^i} \nabla_{\theta^i} \pi_{\theta^i}^i(s, a^i)\Bigr] = 0. \tag{3.6}$$

Thus, replacing $F$ in (3.6) by the value function $V_\theta$ and function $\widetilde{V}_\theta^i$ defined in (3.1) and combined with (3.4), we establish (3.2), which concludes the proof. $\square$

Theorem 3.1 shows that the policy gradient with respect to each $\theta^i$ can be obtained locally using the corresponding score function $\nabla_{\theta^i} \log \pi_{\theta^i}^i$, provided that agent $i$ has an unbiased estimate of the global action-value or advantage functions. However, with only local information, these functions cannot be well estimated since they require the rewards $\{r_t^i\}_{i \in \mathcal{N}}$ of all agents. This motivates our consensus-based MARL algorithms that leverage the communication network to diffuse the local information, which fosters collaboration among the agents.

## 3.1 Algorithms

Now we are ready to develop actor-critic algorithms for networked multi-agent systems. We first propose an algorithm based on the local advantage function $A_\theta^i$ defined in (3.1), which requires estimating the action-value function $Q_\theta$ of policy $\pi_\theta$. More specifically, let $Q(\cdot, \cdot; \omega) \colon \mathcal{S} \times \mathcal{A} \to \mathbb{R}$ be a family of functions parametrized by $\omega \in \mathbb{R}^K$, where $K \ll |\mathcal{S}| \cdot |\mathcal{A}|$. We assume that each agent $i$ maintains its own parameter $\omega^i$ and uses $Q(\cdot, \cdot; \omega^i)$ as a local estimate of $Q_\theta$. Moreover, since $Q_\theta$ in (2.6) involves the globally averaged reward $\bar{r}_t$, to aggregate the local information, we let each agent $i$ share the local parameter $\omega^i$ with its neighbors on the network, in order to reach a consensual estimate of $Q_\theta$. In this way, each agent is able to improve the current policy via the policy gradient theorem.

Specifically, the actor-critic algorithm consists of two steps that proceed at different time scales. In the critic step, the update resembles the action-value TD-learning for policy evaluation in (2.3), followed by a linear combination of its neighbor's parameter estimates. Such a parameter sharing step is also known as the consensus update, which involves a weight matrix $C_t = [c_t(i, j)]_{N \times N}$, where $c_t(i, j)$ is the weight on the message transmitted from $i$ to $j$ at time $t$ The construction of $C_t$ depends on the network topology of $\mathcal{G}_t$; see §4.1 for details. Thus, the critic step iterates as follows

$$\mu_{t+1}^i = (1 - \beta_{\omega,t}) \cdot \mu_t^i + \beta_{\omega,t} \cdot r_{t+1}^i, \quad \widetilde{\omega}_t^i = \omega_t^i + \beta_{\omega,t} \cdot \delta_t^i \cdot \nabla_\omega Q_t(\omega_t^i), \quad \omega_{t+1}^i = \sum_{j \in \mathcal{N}} c_t(i, j) \cdot \widetilde{\omega}_t^j, \tag{3.7}$$

where $\mu_t^i$ tracks the long-term return of agent $i$, $\beta_{\omega,t} > 0$ is the stepsize, and we let $Q_t(\omega) = Q(s_t, a_t; \omega)$ for any $\omega \in \mathbb{R}^K$. Moreover, the local TD-error $\delta_t^i$ in (3.7) is computed as

$$\delta_t^i = r_{t+1}^i - \mu_t^i + Q_{t+1}(\omega_t^i) - Q_t(\omega_t^i). \tag{3.8}$$

As for the actor step, motivated by (3.2) in Theorem 3.1, each agent $i$ improves its policy via

$$\theta_{t+1}^i = \theta_t^i + \beta_{\theta,t} \cdot A_t^i \cdot \psi_t^i, \tag{3.9}$$

**Algorithm 1** The networked actor-critic algorithm based on action-value function
---
**Input:** Initial values of the parameters $\mu_0^i$, $\omega_0^i$, $\widetilde{\omega}_0^i$, $\theta_0^i$, $\forall i \in \mathcal{N}$; the initial state $s_0$ of the MDP, and stepsizes $\{\beta_{\omega,t}\}_{t\geq 0}$ and $\{\beta_{\theta,t}\}_{t\geq 0}$.
Each agent $i \in \mathcal{N}$ executes action $a_0^i \sim \pi_{\theta_0^i}^i(s_0, \cdot)$ and observes joint actions $a_0 = (a_0^1, \ldots, a_0^N)$.
Initialize the iteration counter $t \leftarrow 0$.
**Repeat:**
    **for all** $i \in \mathcal{N}$ **do**
        Observe state $s_{t+1}$, and reward $r_{t+1}^i$.
        Update   $\mu_{t+1}^i \leftarrow (1 - \beta_{\omega,t}) \cdot \mu_t^i + \beta_{\omega,t} \cdot r_{t+1}^i$.
        Select and execute action $a_{t+1}^i \sim \pi_{\theta_t^i}^i(s_{t+1}, \cdot)$.
    **end for**
    Observe joint actions $a_{t+1} = (a_{t+1}^1, \ldots, a_{t+1}^N)$.
    **for all** $i \in \mathcal{N}$ **do**
        Update   $\delta_t^i \leftarrow r_{t+1}^i - \mu_t^i + Q_{t+1}(\omega_t^i) - Q_t(\omega_t^i)$.
        **Critic step:**   $\widetilde{\omega}_t^i \leftarrow \omega_t^i + \beta_{\omega,t} \cdot \delta_t^i \cdot \nabla_\omega Q_t(\omega_t^i)$.
        Update   $A_t^i \leftarrow Q_t(\omega_t^i) - \sum_{a^i \in \mathcal{A}^i} \pi_{\theta_t^i}^i(s_t, a^i) \cdot Q(s_t, a^i, a^{-i}; \omega_t^i)$,    $\psi_t^i \leftarrow \nabla_{\theta^i} \log \pi_{\theta_t^i}^i(s_t, a_t^i)$.
        **Actor step:** $\theta_{t+1}^i \leftarrow \theta_t^i + \beta_{\theta,t} \cdot A_t^i \cdot \psi_t^i$.
        Send $\widetilde{\omega}_t^i$ to the neighbors $\{j \in \mathcal{N} : (i,j) \in \mathcal{E}_t\}$ over the communication network $\mathcal{G}_t$.
    **end for**
    **for all** $i \in \mathcal{N}$ **do**
        **Consensus step:**   $\omega_{t+1}^i \leftarrow \sum_{j \in \mathcal{N}} c_t(i,j) \cdot \widetilde{\omega}_t^j$.
    **end for**
    Update the iteration counter $t \leftarrow t + 1$.
**Until Convergence**

---

where $\beta_{\theta,t} > 0$ is the stepsize. Moreover, $A_t^i$ and $\psi_t^i$ are defined as

$$A_t^i = Q_t(\omega_t^i) - \sum_{a^i \in \mathcal{A}^i} \pi_{\theta_t^i}^i(s_t, a^i) \cdot Q(s_t, a^i, a_t^{-i}; \omega_t^i), \quad \psi_t^i = \nabla_{\theta^i} \log \pi_{\theta_t^i}^i(s_t, a_t^i), \tag{3.10}$$

where the updating rule for $A_t^i$ follows from the definition of $A_\theta^i$ in (3.1).

The update in (3.7) for $\omega_t^i$ resembles the so-termed *diffusion* update in Tu and Sayed (2012); Chen and Sayed (2012) for solving distributed optimization/estimation problems. However, it differs in two main aspects: i) the update direction $\delta_t^i \cdot \nabla_\omega Q_t(\omega_t^i)$ is not the stochastic gradient direction of any well-defined objective function, thus the update is not equivalent to solving any distributed optimization problem; ii) diminishing stepsizes are adopted for possibly almost sure convergence, whereas mean square convergence was established in Tu and Sayed (2012); Chen and Sayed (2012). Thus the proof techniques there do not apply to the analysis of the update in (3.7). We instead resort to the machinery of stochastic approximation for analyzing the convergence of the update under certain assumptions. Moreover, we note that the updates (3.7)-(3.9) preserve the privacy of agents in the sense that no information about the individual reward function or the policy is required for such network-wide collaboration, which inherits one of the advantages of fully decentralized algorithms. We present the steps of this algorithm in Algorithm 1.

For online implementation of Algorithm 1, the joint actions $a_{t+1}$ is needed to evaluate the action-value TD-error $\delta_t^i$. Thus, at time $t$, each agent updates the critic in (3.7) using $(s_t, a_t, s_{t+1}, a_{t+1})$. Moreover, since agent $i$ also needs to store the estimates $\omega_t^i \in \mathbb{R}^K$ and $\theta_t^i \in \mathbb{R}^{m_i}$, the total memory complexity of agent $i$ is $\mathcal{O}(N + m_i + K)$. On the contrary, in the tabular case, each agent $i$ need to maintain a Q-table of dimension $|\mathcal{S}| \cdot |\mathcal{A}| \times |\mathcal{S}| \cdot |\mathcal{A}|$ as in Kar et al. (2013), where $|\mathcal{A}| = \prod_{i \in \mathcal{N}} |A_i|$ grows exponentially with the number of agents $N$ in the system.

Note that Algorithm 1 requires action $a_{t+1}$ to compute the update at time $t$. In the following, we propose an algorithm which only uses the transition at time $t$, namely, the sample $(s_t, a_t, s_{t+1})$, for parameter update. In fact, one can estimate the advantage function $A_\theta$ with the state-value TD-error, since the latter is an unbiased estimate of the former, i.e.,

$$\mathbb{E}\big[\overline{r}_{t+1} - J(\theta) + V_\theta(s_{t+1}) - V_\theta(s_t) \,\big|\, s_t = s, a_t = a, \pi_\theta\big] = A_\theta(s, a), \quad \text{for any } s \in \mathcal{S}, a \in \mathcal{A}. \tag{3.11}$$

To this end, we first estimate $J(\theta)$ and $V_\theta$ with a scalar $\mu$ and a parametrized function $V(\cdot; v) \colon \mathcal{S} \to \mathbb{R}$, respectively, where parameter $v \in \mathbb{R}^L$ with $L \ll |\mathcal{S}|$. Similar to Algorithm 1, each agent $i$ maintains and shares local parameters $\mu^i$ and $v^i$ following the updates

$$\widetilde{\mu}_t^i = (1 - \beta_{v,t}) \cdot \mu_t^i + \beta_{v,t} \cdot r_{t+1}^i, \qquad \mu_{t+1}^i = \sum_{j \in \mathcal{N}} c_t(i, j) \cdot \widetilde{\mu}_t^j, \tag{3.12}$$

$$\delta_t^i = r_{t+1}^i - \mu_t^i + V_{t+1}(v_t^i) - V_t(v_t^i), \qquad \widetilde{v}_t^i = v_t^i + \beta_{v,t} \cdot \delta_t^i \cdot \nabla_v V_t(v_t^i), \qquad v_{t+1}^i = \sum_{j \in \mathcal{N}} c_t(i, j) \cdot \widetilde{v}_t^j, \tag{3.13}$$

where we denote $V_t(v) = V(s_t; v)$ for any $v \in \mathbb{R}^L$, and $\beta_{v,t} > 0$ is the stepsize. With slight abuse of notation, we use $\delta_t^i$ to denote the state-value TD-error of agent $i$. Note that the local $\delta_t^i$ can be used to evaluate the state-value function as for the action-value function evaluation in Algorithm 1. However, it cannot be used directly to estimate the policy gradient following (3.11), since the local $r_{t+1}^i$ is not sampled from the globally averaged reward $\overline{R}$.

Accordingly, we propose to estimate the globally averaged reward function $\overline{R}$ in the critic step as well. Specifically, let $\overline{R}(\cdot, \cdot; \lambda) \colon \mathcal{S} \times \mathcal{A} \to \mathbb{R}$ be the class of parametrized functions, where $\lambda \in \mathbb{R}^M$ is the parameter with $M \ll |\mathcal{S}| \cdot |\mathcal{A}|$. To obtain the estimate $\overline{R}(\cdot, \cdot; \lambda)$ at the faster time scale, we seek to minimize the following weighted mean-square error

$$\underset{\lambda}{\text{minimize}} \quad \sum_{s \in \mathcal{S}, a \in \mathcal{A}} d_\theta(s) \cdot \pi_\theta(s, a) \big[\overline{R}(s, a; \lambda) - \overline{R}(s, a)\big]^2, \tag{3.14}$$

where recall that $\overline{R}(s, a) = \sum_{i \in \mathcal{N}} R^i(s, a) \cdot N^{-1}$ and $d_\theta(s)$ is the stationary distribution of the Markov chain $\{s_t\}_{t \geq 0}$ under policy $\pi_\theta$. The optimization problem (3.14) can be equivalently characterized as

$$\underset{\lambda}{\text{minimize}} \quad \sum_{i \in \mathcal{N}} \sum_{s \in \mathcal{S}, a \in \mathcal{A}} d_\theta(s) \cdot \pi_\theta(s, a) \big[\overline{R}(s, a; \lambda) - R^i(s, a)\big]^2, \tag{3.15}$$

since the two objectives have identical stationary points. The objective (3.15) has the same form of separable objectives over agents as in the distributed optimization literature (Tsitsiklis et al., 1986; Nedic and Ozdaglar, 2009; Boyd et al., 2011; Chen and Sayed, 2012). This connection motivates the following updates for $\lambda_t^i$ to minimize the objective (3.15)

$$\widetilde{\lambda}_t^i = \lambda_t^i + \beta_{v,t} \cdot [r_{t+1}^i - \overline{R}_t(\lambda_t^i)] \cdot \nabla_\lambda \overline{R}_t(\lambda_t^i), \qquad \lambda_{t+1}^i = \sum_{j \in \mathcal{N}} c_t(i, j) \cdot \widetilde{\lambda}_t^j, \tag{3.16}$$

**Algorithm 2** The networked actor-critic algorithm based on state-value TD-error
___
**Input:** Initial values of $\mu_0^i$, $\widetilde{\mu}_0^i$, $v_0^i$, $\widetilde{v}_0^i$, $\lambda_0^i$, $\widetilde{\lambda}_0^i$, $\theta_0^i$, $\forall i \in \mathcal{N}$; the initial state $s_0$ of the MDP, and stepsizes $\{\beta_{v,t}\}_{t\geq 0}$ and $\{\beta_{\theta,t}\}_{t\geq 0}$.
Each agent $i$ implements $a_0^i \sim \pi_{\theta_0^i}(s_0, \cdot)$.
Initialize the step counter $t \leftarrow 0$.
**Repeat:**
  **for all** $i \in \mathcal{N}$ **do**
    Observe state $s_{t+1}$, and reward $r_{t+1}^i$.
    Update   $\widetilde{\mu}_t^i \leftarrow (1-\beta_{v,t}) \cdot \mu_t^i + \beta_{v,t} \cdot r_{t+1}^i$,   $\widetilde{\lambda}_t^i \leftarrow \lambda_t^i + \beta_{v,t} \cdot [r_{t+1}^i - \overline{R}_t(\lambda_t^i)] \cdot \nabla_\lambda \overline{R}_t(\lambda_t^i)$.
    Update   $\delta_t^i \leftarrow r_{t+1}^i - \mu_t^i + V_{t+1}(v_t^i) - V_t(v_t^i)$
    **Critic step:**  $\widetilde{v}_t^i \leftarrow v_t^i + \beta_{v,t} \cdot \delta_t^i \cdot \nabla_v V_t(v_t^i)$.
    Update   $\widetilde{\delta}_t^i \leftarrow \overline{R}_t(\lambda_t^i) - \mu_t^i + V_{t+1}(v_t^i) - V_t(v_t^i)$,   $\psi_t^i \leftarrow \nabla_{\theta^i} \log \pi_{\theta_t^i}^i(s_t, a_t^i)$.
    **Actor step:**  $\theta_{t+1}^i = \theta_t^i + \beta_{\theta,t} \cdot \widetilde{\delta}_t^i \cdot \psi_t^i$.
    Send $\widetilde{\mu}_t^i$, $\widetilde{\lambda}_t^i$, $\widetilde{v}_t^i$ to the neighbors over $\mathcal{G}_t$.
  **end for**
  **for all** $i \in \mathcal{N}$ **do**
    **Consensus step:**  $\mu_{t+1}^i \leftarrow \sum_{j\in\mathcal{N}} c_t(i,j) \cdot \widetilde{\mu}_t^j$,   $\lambda_{t+1}^i \leftarrow \sum_{j\in\mathcal{N}} c_t(i,j) \cdot \widetilde{\lambda}_t^j$,   $v_{t+1}^i \leftarrow \sum_{j\in\mathcal{N}} c_t(i,j) \cdot \widetilde{v}_t^j$.
  **end for**
  Update the iteration counter $t \leftarrow t+1$.
**Until Convergence**
___

where $\overline{R}_t(\lambda) = \overline{R}(s_t, a_t; \lambda)$ for any $\lambda \in \mathbb{R}^M$. The update in (3.16) forms the critic step together with (3.12) and (3.13). We note that the rewards of other agents are not transmitted directly to each agent, and the estimate $\overline{R}(\cdot, \cdot; \lambda)$ can not recover the individual reward function of others. Hence, the consensual estimate of globally averaged reward function does not harm the privacy of agents on their rewards and policies as in Algorithm 1.

Unlike most existing work in distributed optimization, the samples obtained to estimate the gradient of (3.15) are correlated by the Markov chain $\{(s_t, a_t)\}_{t\geq 0}$ under policy $\pi_\theta$. We will also resort to stochastic approximation to analyze the convergence of this update.

The estimate $\overline{R}(\cdot, \cdot; \lambda_t^i)$ is then used to evaluate the globally averaged TD-error $\widetilde{\delta}_t^i$. Accordingly, the actor step motivated by (3.11) becomes

$$\widetilde{\delta}_t^i = \overline{R}_t(\lambda_t^i) - \mu_t^i + V_{t+1}(v_t^i) - V_t(v_t^i), \qquad \theta_{t+1}^i = \theta_t^i + \beta_{\theta,t} \cdot \widetilde{\delta}_t^i \cdot \psi_t^i. \qquad (3.17)$$

where $\beta_{\theta,t} > 0$ is the stepsize and $\psi_t^i$ is as defined in (3.10). The steps of this algorithm are given in Algorithm 2. Similar to Algorithm 1, the online implementation of Algorithm 2 requires the memory complexity of $\mathcal{O}(N + L + M + m_i)$ for each agent $i$, which results in a great reduction in contrast to the tabular case when $N$ is large.

Note that both algorithms are applicable to general function approximators including deep neural networks. In addition, when linear approximation is applied, we provide convergence guarantees in §4.

# 4 Theoretical Results

In this section, we establish theoretical results for the proposed algorithms. Specifically, with linear function approximation, we show the convergence of both Algorithms 1 and 2, with complete proofs relegated to the next section. We start with the following three assumptions which apply to both algorithms.

**Assumption 4.1.** The update of the policy parameter $\theta_t^i$ includes a local projection operator, $\Gamma^i : \mathbb{R}^{m_i} \to \Theta^i \subset \mathbb{R}^{m_i}$, that projects any $\theta_t^i$ onto a compact set $\Theta^i$. Also, we assume that $\Theta = \prod_{i=1}^N \Theta^i$ is large enough to include at least one local minimum of $J(\theta)$.

This projection is a common technique for stabilizing the transient behavior of stochastic approximation algorithms (Kushner and Yin, 2003). It has been a standard assumption in many analyses for two-time-scale reinforcement learning algorithms (Abounadi et al., 2001; Bhatnagar et al., 2009; Degris et al., 2012; Prasad et al., 2014). Note that this projection is merely for analysis purposes and may not be necessary when updating $\theta^i$ in experiments, as we will illustrate later.

**Assumption 4.2.** The instantaneous reward $r_t^i$ is uniformly bounded for any $i \in \mathcal{N}$ and $t \geq 0$.

We remark that the boundedness assumption on the rewards is rather mild, which is readily satisfied in practical MDP models with finite state and action spaces.

Furthermore, we make the following assumption on the weight matrix $\{C_t\}_{t \geq 0}$ for the consensus updates in both algorithms.

**Assumption 4.3.** The sequence of nonnegative random matrices $\{C_t\}_{t \geq 0} \subseteq \mathbb{R}^{N \times N}$ satisfies

(a.1) $C_t$ is row stochastic and $\mathbb{E}(C_t)$ is column stochastic. That is, $C_t \mathbb{1} = \mathbb{1}$ and $\mathbb{1}^\top \mathbb{E}(C_t) = \mathbb{1}^\top$. Furthermore, there exists a constant $\eta \in (0,1)$ such that, for any $c_t(i,j) > 0$, we have $c_t(i,j) \geq \eta$.

(a.2) $C_t$ respects the communication graph $\mathcal{G}_t$, i.e., $c_t(i,j) = 0$ if $(i,j) \notin \mathcal{E}_t$.

(a.3) The spectral norm of $\mathbb{E}[C_t^\top \cdot (I - \mathbb{1}\mathbb{1}^\top/N) \cdot C_t]$ is strictly smaller than one.

(a.4) Given the $\sigma$-algebra generated by the random variables before time $t$, $C_t$ is conditionally independent of $r_{t+1}^i$ for any $i \in \mathcal{N}$.

We take the matrix $C_t$ to be random for the sake of generality. The randomness can be attributed to either the randomness of the time-varying graph $\mathcal{G}_t$, e.g., random link failures in sensor networks (Kar and Moura, 2010), or the randomness of the consensus algorithms, e.g., randomized gossip schemes (Boyd et al., 2006). The condition (a.1) is standard in developing consensus algorithms, which guarantees convergence of the update for each agent to a common vector. As noted in Bianchi et al. (2013), the matrix $C_t$ here is only required to be column stochastic in mean, as opposed to most other gossip or consensus algorithms that require $C_t$ to be doubly stochastic[2]. Note that row stochasticity constraint $C_t \mathbb{1} = \mathbb{1}$ is local, since it simply requires each agent to make the weights assigned to the updates coming from its neighbors summing to one. The lower boundedness of the weights is needed to ensure that the limit $\lim_t C_t C_{t-1} \cdots C_0$ exists (Nedic and Ozdaglar, 2009), which is required in the proof of the stability of the consensus update (see Appendix §A). The conditions (a.2) and (a.3) are related to the connectivity of the communication network $\mathcal{G}_t$. It follows from Boyd et al. (2006); Aysal et al. (2009) that for the gossip type of communication schemes, (a.3) holds if and only if the underlying communication graph is connected. See Nedich et al. (2016) for more discussion on the connection between the spectrum property (a.3) and the connectivity of time-varying graphs. The condition (a.4) means that $C_t$ and $r_{t+1}$ are independent conditioned on the past. This is common for practical multi-agent systems, since the random communication link failures and the gossip schemes are usually independent of the past history and irrelevant to the rewards received by the agents.

---
[2]A stochastic matrix $P$ is doubly stochastic if it is both row and column stochastic.

One particular choice of the weights in $C_t$ that relies on only local information of the agents is known as Metropolis weights (Xiao et al., 2005; Cattivelli et al., 2008)

$$c_t(i,j) = \left\{1 + \max[d_t(i), d_t(j)]\right\}^{-1}, \ \forall (i,j) \in \mathcal{E}_t, \qquad c_t(i,i) = 1 - \sum_{j \in \mathcal{N}_t(i)} c_t(i,j), \ \forall i \in \mathcal{N},$$

where $\mathcal{N}_t(i) = \{j \in \mathcal{N} : (i,j) \in \mathcal{E}_t\}$ is the set of neighbors of agent $i$ at time $t$, and $d_t(i) = |\mathcal{N}_t(i)|$ is the degree of agent $i$. Other common choices of $C_t$ that satisfy the conditions (a.1)-(a.3) in Assumption 4.3 include pairwise gossip (Boyd et al., 2006), broadcast gossip (Aysal et al., 2009), and network dropouts (Bianchi et al., 2013).

### 4.1 Convergence of Algorithm 1

To show the convergence of Algorithm 1, we make the following additional assumptions on the value functions and the stepsizes.

**Assumption 4.4.** For each agent $i$, the action-value function is parametrized by the class of linear functions, i.e., $Q(s,a;\omega) = \omega^\top \phi(s,a)$ where $\phi(s,a) = [\phi_1(s,a), \cdots, \phi_K(s,a)]^\top \in \mathbb{R}^K$ is the feature associated with the state-action pair $(s,a)$. The feature vectors $\phi(s,a)$ are uniformly bounded for any $s \in \mathcal{S}, a \in \mathcal{A}$. Furthermore, the feature matrix $\Phi \in \mathbb{R}^{|\mathcal{S}|\cdot|\mathcal{A}| \times K}$ has full column rank, where the $k$-th column of $\Phi$ is $[\phi_k(s,a), s \in \mathcal{S}, a \in \mathcal{A}]^\top$ for any $k \in [K]$. Also, for any $u \in \mathbb{R}^K$, $\Phi u \neq \mathbb{1}$.

We focus here on the version of Algorithm 1 with linear function approximation for the purpose of theoretical analysis. Note that the TD-learning-based policy evaluation with nonlinear function approximation may fail to converge (Tsitsiklis and Van Roy, 1997). Thus, the convergence of AC algorithms with nonlinear function approximation is not clear even for the single-agent setting. To the best of our knowledge, all existing AC algorithms with theoretical guarantees are built upon linear function approximation, e.g., Konda and Tsitsiklis (2000); Bhatnagar et al. (2009); Bhatnagar (2010). The assumption on the feature matrix $\Phi$ is standard and has also been made in Tsitsiklis and Van Roy (1997, 1999); Bhatnagar et al. (2009) to ensure that policy evaluation has a unique asymptotically stable solution, as we will show in the proof.

We then make the assumption that the stepsizes $\beta_{\omega,t}$ and $\beta_{\theta,t}$ satisfy the following standard condition similar to that for single-agent AC algorithms with two-time-scale updates.

**Assumption 4.5.** The stepsizes $\beta_{\omega,t}$ and $\beta_{\theta,t}$ satisfy

$$\sum_t \beta_{\omega,t} = \sum_t \beta_{\theta,t} = \infty, \quad \sum_t \beta_{\omega,t}^2 + \beta_{\theta,t}^2 < \infty, \quad \beta_{\theta,t} = o(\beta_{\omega,t}).$$

In addition, $\lim_{t \to \infty} \beta_{\omega,t+1} \cdot \beta_{\omega,t}^{-1} = 1$.

For the convergence analysis of Algorithm 1, we use the two-time-scale stochastic approximation (SA) technique (Borkar, 2008). In particular, we first analyze the convergence of the critic step at the faster time scale, where the joint policy $\pi_\theta$ is assumed to be fixed. Then we analyze the convergence of the policy parameter $\theta_t$ upon the convergence of the critic step.

For notational simplicity, we define $P^\theta(s',a'|s,a) = P(s'|s,a)\pi_\theta(s',a')$[3] and $D_\theta^{s,a} = \text{diag}[d_\theta(s) \cdot \pi_\theta(s,a), s \in \mathcal{S}, a \in \mathcal{A}]$. Recall that $\overline{R} = [\overline{R}(s,a), s \in \mathcal{S}, a \in \mathcal{A}]^\top \in \mathbb{R}^{|\mathcal{S}| \cdot |\mathcal{A}|}$ has element $\overline{R}(s,a) = \sum_{i \in \mathcal{N}} R^i(s,a) \cdot N^{-1}$. Then we define the operator $T_\theta^Q : \mathbb{R}^{|\mathcal{S}| \cdot |\mathcal{A}|} \to \mathbb{R}^{|\mathcal{S}| \cdot |\mathcal{A}|}$ for any action-value vector $Q \in \mathbb{R}^{|\mathcal{S}| \cdot |\mathcal{A}|}$ as

$$T_\theta^Q(Q) = \overline{R} - J(\theta) \cdot \mathbb{1} + P^\theta Q. \tag{4.1}$$

With these notations, we establish the convergence of the critic step (3.7) and (3.8) as follows.

**Theorem 4.6.** *Under Assumptions 2.2, and 4.2-4.5, for any given policy $\pi_\theta$, with $\{\mu_t^i\}$ and $\{\omega_t^i\}$ generated from (3.7) and (3.8), we have $\lim_t \sum_{i \in \mathcal{N}} \mu_t^i \cdot N^{-1} = J(\theta)$ and $\lim_t \omega_t^i = \omega_\theta$ almost surely (a.s.) for any $i \in \mathcal{N}$, where*

$$J(\theta) = \sum_{s \in \mathcal{S}} d_\theta(s) \sum_{a \in \mathcal{A}} \pi_\theta(s,a) \overline{R}(s,a)$$

*is the globally long-term averaged return under joint policy $\pi_\theta$, and $\omega_\theta$ is the unique solution to*

$$\Phi^\top D_\theta^{s,a} \left[ T_\theta^Q(\Phi \omega_\theta) - \Phi \omega_\theta \right] = 0. \tag{4.2}$$

We note that the solution to (4.2) is the limiting point of the TD(0) algorithm (Tsitsiklis and Van Roy, 1999), except that here we approximate the action-value function $Q_\theta$ rather than the state-value function $V_\theta$. This point is also the minimizer of the Mean Square Projected Bellman Error (MSPBE) (Dann et al., 2014), i.e., the solution to

$$\underset{\omega}{\text{minimize}} \quad \left\| \Phi \omega - \Pi T_\theta^Q(\Phi \omega) \right\|_{D_\theta^{s,a}}^2, \tag{4.3}$$

where $\Pi$ is the operator that projects a vector to the space spanned by the columns of $\Phi$, and $\|\cdot\|_{D_\theta^{s,a}}^2$ denotes the Euclidean norm weighted by the matrix $D_\theta^{s,a}$. Thus, Theorem 4.6 states that each agent is enabled to obtain a copy of the approximation of the *globally* averaged action-value function, i.e., $\omega_t^i \to \omega_\theta$ for all $i \in \mathcal{N}$, even with only local rewards and information exchange with neighbors. This approximation of global value function is then adopted in the actor step to estimate the policy gradient for each agent.

To show convergence of the actor step, we define the quantities $A_{t,\theta}^i$ and $\psi_{t,\theta}^i$ as

$$A_{t,\theta}^i = \phi_t^\top \omega_\theta - \sum_{a^i \in \mathcal{A}^i} \pi_{\theta^i}^i(s_t, a^i) \cdot \phi(s_t, a^i, a_t^{-i})^\top \omega_\theta, \qquad \psi_{t,\theta}^i = \nabla_{\theta^i} \log \pi_{\theta^i}^i(s_t, a_t^i), \tag{4.4}$$

where we denote $\phi_t = \phi(s_t, a_t)$. Recall that $\Gamma^i$ is the operator that projects any vector onto the compact set $\Theta^i$ (see Assumption 4.1). Define a vector $\hat{\Gamma}^i(\cdot)$ as

$$\hat{\Gamma}^i[g(\theta)] = \lim_{0 < \eta \to 0} \left\{ \Gamma^i[\theta^i + \eta \cdot g(\theta)] - \theta^i \right\} / \eta \tag{4.5}$$

for any $\theta \in \Theta$ and $g : \Theta \to \mathbb{R}^{\sum_{i \in \mathcal{N}} m_i}$ a continuous function. In case the limit above is not unique, let $\hat{\Gamma}^i[g(\theta)]$ be the set of all possible limit points of (4.5). Then we state the convergence of Algorithm 1 with linear function approximation as follows.

---

[3]With slight abuse of notation, the expression $P^\theta$ has the same form as the transition probability matrix of the Markov chain $\{s_t\}_{t \geq 0}$ under policy $\pi_\theta$ (see the definition in (2.4)). These two matrices can be easily differentiated by the context.

**Theorem 4.7.** Under Assumptions 2.2 and 4.1-4.5, the policy parameter $\theta_t^i$ obtained from (3.9) converges almost surely to a point in the set of asymptotically stable equilibria of

$$\dot{\theta}^i = \hat{\Gamma}^i \Big[ \mathbb{E}_{s_t \sim d_\theta, a_t \sim \pi_\theta} \Big( A_{t,\theta}^i \cdot \psi_{t,\theta}^i \Big) \Big], \quad \text{for any } i \in \mathcal{N}. \tag{4.6}$$

Note that Theorem 4.7 is in the same spirit as the convergence results for single-agent AC algorithms with linear function approximation (Bhatnagar et al., 2009; Bhatnagar, 2010). Since the linear features here are not restricted to *compatible* features as in Tsitsiklis and Van Roy (1997); Sutton et al. (2000), convergence to the stationary point of $\mathbb{E}_{s_t \sim d_\theta, a_t \sim \pi_\theta} \Big( A_{t,\theta}^i \cdot \psi_{t,\theta}^i \Big) = 0$ in the set $\Theta^i$ is the best one can hope for any AC algorithms with general linear function approximators, even for the single-agent setting (Bhatnagar et al., 2009; Degris et al., 2012).

In general, since the linear function approximator has nonzero approximation error for $Q_\theta$, even the convergent term $A_{t,\theta}^i \cdot \psi_{t,\theta}^i$ from the critic step may not be an unbiased estimate of $\nabla_{\theta^i} J(\theta)$. In particular, the estimate of policy gradient $A_{t,\theta}^i \cdot \psi_t^i$ satisfies

$$\mathbb{E}_{s_t \sim d_\theta, a_t \sim \pi_\theta} \Big( A_{t,\theta}^i \cdot \psi_{t,\theta}^i \Big) = \nabla_{\theta^i} J(\theta) + \mathbb{E}_{s_t \sim d_\theta, a_t \sim \pi_\theta} \Big\{ \Big[ \phi_t^\top \omega_\theta - Q_\theta(s_t, a_t) \Big] \cdot \psi_{t,\theta}^i \Big\}.$$

Thus, the convergent point of (4.6) corresponds to a small neighborhood of a local optimum of $J(\theta)$, i.e, $\nabla_{\theta^i} J(\theta) = 0$, provided the error term from the approximation error for the action-value function $\phi_t^\top \omega_\theta - Q_\theta(s_t, a_t)$ is small.

## 4.2 Convergence of Algorithm 2

Similar to Algorithm 1, we need two additional assumptions for the convergence of Algorithm 2.

**Assumption 4.8.** For each agent $i$, the state-value function and the globally averaged reward function are both parametrized by the class of linear functions, i.e., $V(s;v) = v^\top \varphi(s)$ and $\overline{R}(s,a;\lambda) = \lambda^\top f(s,a)$, where $\varphi(s) = [\varphi_1(s), \cdots, \varphi_K(s)]^\top \in \mathbb{R}^L$ and $f(s,a) = [f_1(s,a), \cdots, f_M(s,a)]^\top \in \mathbb{R}^M$ are the features associated with $s$ and $(s,a)$, respectively. The feature vectors $\varphi(s)$ and $f(s,a)$ are uniformly bounded for any $s \in \mathcal{S}, a \in \mathcal{A}$. Furthermore, let the feature matrix $\Phi \in \mathbb{R}^{|\mathcal{S}| \times L}$ have $[\varphi_\ell(s), s \in \mathcal{S}]^\top$ as its $\ell$-th column for any $\ell \in [L]$, and the feature matrix $F \in \mathbb{R}^{|\mathcal{S}| \cdot |\mathcal{A}| \times M}$ have $[f_m(s,a), s \in \mathcal{S}, a \in \mathcal{A}]^\top$ as its $m$-th column for any $m \in [M]$. Then, both $\Phi$ and $F$ have full column rank, and for any $u \in \mathbb{R}^L$, $\Phi u \neq \mathbb{1}$.

For convergence analysis, we focus on Algorithm 2 with linear function approximation for both the state-value function and the globally averaged reward function.

**Assumption 4.9.** The stepsizes $\beta_{v,t}$ and $\beta_{\theta,t}$ satisfy

$$\sum_t \beta_{v,t} = \sum_t \beta_{\theta,t} = \infty, \quad \sum_t \beta_{v,t}^2 + \beta_{\theta,t}^2 < \infty, \quad \beta_{\theta,t} = o(\beta_{v,t}).$$

In addition, $\lim_{t \to \infty} \beta_{v,t+1} \cdot \beta_{v,t}^{-1} = 1$.

Recall $P^\theta(s'|s) = \sum_{a \in \mathcal{A}} P(s'|s,a) \pi_\theta(s,a)$ defined in (2.4) and $d_\theta(s)$ denote, respectively, the transition probability and stationary distribution of the Markov chain $\{s_t\}_{t \geq 0}$ under policy $\pi_\theta$. Let $D_\theta^s = \text{diag}[d_\theta(s), s \in \mathcal{S}]$. Also recall that $\overline{R}_\theta = [\overline{R}_\theta(s), s \in \mathcal{S}]^\top \in \mathbb{R}^{|\mathcal{S}|}$ has element $\overline{R}_\theta(s) = \sum_a \pi_\theta(s,a)\overline{R}(s,a)$. We thus define the operator $T_\theta^V : \mathbb{R}^{|\mathcal{S}|} \to \mathbb{R}^{|\mathcal{S}|}$ for any state-value vector $V \in \mathbb{R}^{|\mathcal{S}|}$ as

$$T_\theta^V(V) = \overline{R}_\theta - J(\theta) \cdot \mathbb{1} + P^\theta V. \tag{4.7}$$

We now state the convergence of the critic step (3.12), (3.13), and (3.16) as follows.

**Theorem 4.10.** Under Assumptions 2.2, 4.2, 4.3, 4.8, and 4.9, for any given policy $\pi_\theta$, with $\{\lambda_t^i\}$, $\{\mu_t^i\}$, and $\{v_t^i\}$ generated from (3.12), (3.13), and (3.16), we have $\lim_t \mu_t^i = J(\theta)$, $\lim_t \lambda_t^i = \lambda_\theta$, and $\lim_t v_t^i = v_\theta$ a.s. for any $i \in \mathcal{N}$, where $J(\theta)$ is the globally averaged return under joint policy $\pi_\theta$, $\lambda_\theta$ and $v_\theta$ are the unique solutions to

$$F^\top D_\theta^{s,a}(\overline{R} - F\lambda_\theta) = 0, \qquad \Phi^\top D_\theta^s \big[T_\theta^V(\Phi v_\theta) - \Phi v_\theta\big] = 0, \tag{4.8}$$

respectively, where we have $D_\theta^{s,a} = \text{diag}[d_\theta(s) \cdot \pi_\theta(s,a), s \in \mathcal{S}, a \in \mathcal{A}]$ and $D_\theta^s = \text{diag}[d_\theta(s), s \in \mathcal{S}]$.

Similarly, the solution $v_\theta$ in (4.8) is exactly the limiting point of the TD(0) algorithm as if the rewards of all others are observable to each agent. Moreover, the solution $\lambda_\theta$ in (4.8) corresponds to the unique minimizer of the problem (3.14) under Assumption 4.8. Both $v_\theta$ and $\lambda_\theta$ are used to define the TD-error $\widetilde{\delta}_{t,\theta}^i$ upon the convergence of the critic step in Algorithm 2, notably,

$$\widetilde{\delta}_{t,\theta}^i = f_t^\top \lambda_\theta - J(\theta) + \varphi_{t+1}^\top v_\theta - \varphi_t^\top v_\theta, \tag{4.9}$$

where we define $f_t = f(s_t, a_t)$ and $\varphi_t = \varphi(s_t)$. Recalling that $\psi_{t,\theta}^i = \nabla_{\theta^i} \log \pi_{\theta^i}^i(s_t, a_t^i)$, we have the following theorem on the convergence of Algorithm 2 with linear function approximation.

**Theorem 4.11.** Under Assumptions 2.2, 4.1-4.3, 4.8, and 4.9, the policy parameter $\theta_t^i$ obtained from (3.17) converges almost surely to a point in the set of asymptotically stable equilibria of

$$\dot{\theta}^i = \hat{\Gamma}^i \Big[\mathbb{E}_{s_t \sim d_\theta, a_t \sim \pi_\theta}\big(\widetilde{\delta}_{t,\theta}^i \cdot \psi_{t,\theta}^i\big)\Big], \quad \text{for any } i \in \mathcal{N}. \tag{4.10}$$

Note that $\widetilde{\delta}_{t,\theta}^i \cdot \psi_{t,\theta}^i$ may not be an unbiased estimate of $\nabla_{\theta^i} J(\theta)$. In particular, we have

$$\mathbb{E}_{s_t \sim d_\theta, a_t \sim \pi_\theta}\big(\widetilde{\delta}_{t,\theta}^i \cdot \psi_{t,\theta}^i\big) = \nabla_{\theta^i} J(\theta) + \mathbb{E}_{s_t \sim d_\theta, a_t \sim \pi_\theta}\Big\{\big[f_t^\top \lambda_\theta - \overline{R}(s_t, a_t)\big]\psi_{t,\theta}^i\Big\} + \mathbb{E}_{s_t \sim d_\theta}\Big\{\big[\varphi_t^\top v_\theta - V_\theta(s_t)\big]\psi_{t,\theta}^i\Big\}.$$

If the approximation errors of both functions $f_t^\top \lambda_\theta - \overline{R}(s_t, a_t)$ and $\varphi_t^\top v_\theta - V_\theta(s_t)$ are small, the convergent point of (4.10) is close to the local optimum of $J(\theta)$.

# 5 Proofs of the Main Results

In this section, we provide the proofs of the convergence results in Section §4. We first provide a detailed proof for the convergence of Algorithm 1, and then prove the two theorems related to Algorithm 2 by drawing parallels with the first two proofs.

## 5.1 Proof of Theorem 4.6

To proceed with the proof, we first establish the stability of the update $\{\omega_t\}$. This stability condition serves as an assumption in the original two-time-scale SA analysis (Borkar, 2008, Chapter 6.1). It is usually verified separately using some other sufficient conditions (Borkar and Meyn, 2000; Andrieu et al., 2005). We will directly use the lemma in the convergence analysis to follow and defer its proof to Appendix §A.

**Lemma 5.1.** Under Assumptions 2.2, and 4.2-4.5, the sequence $\{\omega_t^i\}$ generated from (3.7) is bounded almost surely, i.e., $\sup_t \|\omega_t^i\| < \infty$ a.s., for any $i \in \mathcal{N}$.

As in the classical two-time-scale SA analysis (Borkar, 2008), we let the policy parameter $\theta_t$ to be fixed as $\theta_t \equiv \theta$ when analyzing the convergence of the critic step. This allows us to show that $\omega_t$ will converge to some $\omega_\theta$ depending on $\theta$, which can be further utilized to simplify the proof of convergence for the slow time scale. In fact, with linear function approximation, one can rewrite the actor step (3.9) for Algorithm 1 as

$$\theta_{t+1}^i = \Gamma^i \left( \theta_t^i + \beta_{\omega,t} \cdot \frac{\beta_{\theta,t}}{\beta_{\omega,t}} \cdot A_t^i \cdot \psi_t^i \right), \tag{5.1}$$

where the projection $\Gamma^i$ follows from Assumption 4.1. Note that $A_t^i$ is also bounded a.s., since the parameter $\omega_t^i$ is bounded by Lemma 5.1, and the feature $\phi_t$ is bounded by and Assumption 4.4. Moreover, $\psi_t^i$ is bounded a.s. by Assumption 2.2 since it is a continuous function over a bounded set $\Theta^i$. Therefore, $\sup_t \|A_t^i \cdot \psi_t^i\| < \infty$ a.s. Now since $\beta_{\theta,t} \cdot \beta_{\omega,t}^{-1} \to 0$ by Assumption 4.5, it follows that $\beta_{\theta,t} \cdot \beta_{\omega,t}^{-1} \cdot A_t^i \cdot \psi_t^i \to 0$ as $t \to \infty$. Now the update in (5.1) can be viewed to track the ordinary differential equation (ODE) $\dot{\theta}^i(t) = 0$. Hence, one may let $\theta_t$ be a constant when analyzing the faster update of $\omega_t^i$. For notational simplicity, we eliminate the notations associated with $\theta$ unless otherwise noted.

Let $\{\mathcal{F}_{t,1}\}$ be the filtration with $\mathcal{F}_{t,1} = \sigma(r_\tau, \mu_\tau, \omega_\tau, s_\tau, a_\tau, C_{\tau-1}, \tau \leq t)$, which is an increasing $\sigma$-algebra over time $t$. For notational convenience, let $r_t = (r_t^1, \cdots, r_t^N)^\top$, $\mu_t = (\mu_t^1, \cdots, \mu_t^N)^\top$, $\omega_t = [(\omega_t^1)^\top, \cdots, (\omega_t^N)^\top]^\top$, and $\delta_t = [(\delta_t^1)^\top, \cdots, (\delta_t^N)^\top]^\top$. The update of $\omega_t$ in (3.7) can be rewritten in a compact form as

$$\omega_{t+1} = (C_t \otimes I)(\omega_t + \beta_{\omega,t} \cdot y_{t+1}), \tag{5.2}$$

where $\otimes$ denotes the Kronecker product, $I \in \mathbb{R}^{K \times K}$ is the identity matrix, and $y_{t+1} = (\delta_t^1 \phi_t^\top, \cdots, \delta_t^N \phi_t^\top)^\top \in \mathbb{R}^{KN}$. Define the operator $\langle \cdot \rangle : \mathbb{R}^{KN} \to \mathbb{R}^K$ by letting

$$\langle \omega \rangle = \frac{1}{N}(\mathbf{1}^\top \otimes I)\omega = \frac{1}{N} \sum_{i \in \mathcal{N}} \omega^i \tag{5.3}$$

for any $\omega = [(\omega^1)^\top, \cdots, (\omega^N)^\top]^\top \in \mathbb{R}^{KN}$ and $\omega^i \in \mathbb{R}^K$ with $i \in \mathcal{N}$. That is, $\langle \omega \rangle \in \mathbb{R}^K$ represents the average of the vectors in $\{\omega^1, \cdots, \omega^N\}$, which are local to individual agents. Let $\mathcal{J} = (1/N \cdot \mathbf{1}\mathbf{1}^\top) \otimes I$ be the projection operator that projects the vector into the *consensus* subspace $\{\mathbf{1} \otimes u : u \in \mathbb{R}^K\}$. Thus we have $\mathcal{J}\omega = \mathbf{1} \otimes \langle \omega \rangle$. Moreover, we define $\mathcal{J}_\perp$ as the operator that projects the vector to the *disagreement* subspace, i.e., $\mathcal{J}_\perp = I - \mathcal{J}$. Thus the disagreement vector $\omega_\perp = \mathcal{J}_\perp \omega$ is written as

$$\omega_\perp = \mathcal{J}_\perp \omega = \omega - \mathbf{1} \otimes \langle \omega \rangle. \tag{5.4}$$

The proof of Theorem 4.6 then consists of two steps. In particular, we separate the iteration $\omega_t$ as the sum of a vector in this *consensus* space and a vector in the *disagreement* space, i.e., $\omega_t = \omega_{\perp,t} + \mathbf{1} \otimes \langle \omega_t \rangle$. We first show the a.s. convergence of the disagreement vector sequence $\{\omega_{\perp,t}\}$ to zero. Then, we prove that the consensus vector sequence $\{\mathbf{1} \otimes \langle \omega_t \rangle\}$ converges to the equilibrium such that $\langle \omega_t \rangle$ satisfies (4.2).

**Step 1.** In this step, we establish that $\lim_t \omega_{\perp,t} = 0$ a.s. To this end, we first have the following lemma on the boundedness of the sequence $\{\mu_t^i\}$ for any $i \in \mathcal{N}$.

**Lemma 5.2.** *Under Assumptions 2.2 and 4.2, the sequence $\{\mu_t^i\}$ generated as in (3.7) is bounded almost surely, i.e., $\sup_t |\mu_t^i| < \infty$ a.s., for any $i \in \mathcal{N}$.*

*Proof.* The local update in (3.7) forms a stochastic approximation iteration, whose asymptotic behavior can be captured by the ODE

$$\dot{\mu}^i = -\mu^i + \sum_{s \in \mathcal{S}} d_\theta(s) \sum_{a \in \mathcal{A}} \pi_\theta(s,a) R^i(s,a). \qquad (5.5)$$

Let $f(\mu^i)$ denote the right hand side (RHS) of (5.5), which is Lipschitz continuous in $\mu^i$. Moreover, define $f_c(\mu^i) = f(c\mu^i) \cdot c^{-1}$, then $f_\infty(\mu^i) = \lim_c f(c\mu^i) \cdot c^{-1} = -\mu^i$ exists. Therefore, the ODE $\dot{\mu}^i = f_\infty(\mu^i)$ has origin as the unique asymptotically stable equilibrium. In addition, since $r_t^i$ is uniformly bounded, we have

$$\mathbb{E}\big[\big|r_{t+1}^i - \mathbb{E}(r_{t+1}^i \,|\, \mathcal{F}_{t,1})\big|^2 \,\big|\, \mathcal{F}_{t,1}\big] \leq K_0 \cdot (1 + |\mu_t^i|^2)$$

for some $K_0 < \infty$. Therefore, the conditions (a.1) and (a.4) in Assumption B.1 are satisfied. (See Appendix §B.1 for details.) By Assumption 2.2, (a.2) in Assumption B.1 also holds. We thus conclude that $\sup_t |\mu_t^i| < \infty$ from Theorem B.3 (see also Theorem 9 on page 74-75 in Borkar (2008)). $\square$

Let $z_t^i = [\mu_t^i, (\omega_t^i)^\top]^\top$ and $z_t = [(z_t^1)^\top, \cdots, (z_t^N)^\top]^\top$. By Lemma 5.1, we have $\mathbb{P}(\sup_t \|z_t\| < \infty) = 1$, which means that $\mathbb{P}(\bigcup_{M \in \mathbb{Z}^+} \{\sup_t \|z_t\| \leq M\}) = 1$, with $\mathbb{Z}^+$ denoting the set of positive integers. Hence, it suffices to show that $\lim_t \omega_{\perp,t} \mathbb{I}_{\{\sup_t \|z_t\| \leq M\}} = 0$, for any $M \in \mathbb{Z}^+$, where $\mathbb{I}_{\{\cdot\}}$ is the indicator function. We then establish that $\mathbb{E}(\|\beta_{\omega,t}^{-1} \omega_{\perp,t}\|^2)$ is bounded on $\{\sup_t \|z_t\| \leq M\}$, for any $M > 0$.

**Lemma 5.3.** *Under Assumptions 4.2-4.5, for any $M > 0$, we have*

$$\sup_t \mathbb{E}\big(\|\beta_{\omega,t}^{-1} \omega_{\perp,t}\|^2 \mathbb{I}_{\{\sup_t \|z_t\| \leq M\}}\big) < \infty.$$

*Proof.* First note that by (a.1) in Assumption 4.3 and the fact that $(A \otimes B)(C \otimes D) = (AC) \otimes (BD)$, we have

$$(C_t \otimes I)(\mathbb{1} \otimes \langle \omega \rangle) = (C_t \mathbb{1}) \otimes \langle \omega \rangle = \mathbb{1} \otimes \langle \omega \rangle.$$

Hence, $\omega_{\perp,t+1}$ has the form $\omega_{\perp,t+1} = \mathcal{J}_\perp[(C_t \otimes I)(\omega_t + \beta_{\omega,t} y_{t+1})] = \mathcal{J}_\perp[(C_t \otimes I)(\omega_{\perp,t} + \beta_{\omega,t} y_{t+1})]$, since $\mathcal{J}_\perp(\mathbb{1} \otimes \langle \omega \rangle)$ is zero. Thus, by the definition of $\mathcal{J}_\perp$ in (5.4), the vector $\omega_{\perp,t+1}$ satisfies

$$\omega_{\perp,t+1} = [(I - \mathbb{1}\mathbb{1}^\top/N) \otimes I](C_t \otimes I)(\omega_{\perp,t} + \beta_{\omega,t} y_{t+1}) = [(I - \mathbb{1}\mathbb{1}^\top/N)C_t \otimes I](\omega_{\perp,t} + \beta_{\omega,t} y_{t+1}). \qquad (5.6)$$

Thus, we have

$$\mathbb{E}\big(\big\|\beta_{\omega,t+1}^{-1} \omega_{\perp,t+1}\big\|^2 \,\big|\, \mathcal{F}_{t,1}\big) = \frac{\beta_{\omega,t}^2}{\beta_{\omega,t+1}^2} \cdot \mathbb{E}\big\{\big(\beta_{\omega,t}^{-1} \omega_{\perp,t} + y_{t+1}\big)^\top \big[C_t^\top (I - \mathbb{1}\mathbb{1}^\top/N) C_t \otimes I\big] \big(\beta_{\omega,t}^{-1} \omega_{\perp,t} + y_{t+1}\big) \,\big|\, \mathcal{F}_{t,1}\big\}$$

$$\leq \frac{\beta_{\omega,t}^2}{\beta_{\omega,t+1}^2} \cdot \rho \cdot \mathbb{E}\big[\big(\beta_{\omega,t}^{-1} \omega_{\perp,t} + y_{t+1}\big)^\top \big(\beta_{\omega,t}^{-1} \omega_{\perp,t} + y_{t+1}\big) \,\big|\, \mathcal{F}_{t,1}\big]$$

$$\leq \frac{\beta_{\omega,t}^2}{\beta_{\omega,t+1}^2} \cdot \rho \cdot \Big\{\big\|\beta_{\omega,t}^{-1} \omega_{\perp,t}\big\|^2 + 2 \cdot \big\|\beta_{\omega,t}^{-1} \omega_{\perp,t}\big\| \cdot \big[\mathbb{E}\big(\|y_{t+1}\|^2 \,\big|\, \mathcal{F}_{t,1}\big)\big]^{\frac{1}{2}} + \mathbb{E}\big(\|y_{t+1}\|^2 \,\big|\, \mathcal{F}_{t,1}\big)\Big\}, \qquad (5.7)$$

where $\rho$ represents the spectral norm of $\mathbb{E}[C_t^\top(I - \mathbb{1}\mathbb{1}^\top/N)C_t]$. By (a.3) in Assumption 4.3, we have $\rho \in [0, 1)$. The first inequality in (5.7) is due to the conditional independence of $C_t$ and $r_{t+1}^i$ for all $i \in \mathcal{N}$, and thus $y_{t+1}$, by (a.4) in Assumption 4.3, and the second inequality is due to the Cauchy-Schwarz inequality. Moreover, by the definition of $y_{t+1}$, we have

$$\mathbb{E}(\|y_{t+1}\|^2 | \mathcal{F}_{t,1}) = \mathbb{E}\left(\sum_{i\in\mathcal{N}}\|\delta_t^i \phi_t\|^2 \Big| \mathcal{F}_{t,1}\right) = \mathbb{E}\left[\sum_{i\in\mathcal{N}}\|(r_{t+1}^i - \mu_t^i + \phi_{t+1}^\top \omega_t^i - \phi_t^\top \omega_t^i)\phi_t\|^2 \Big| \mathcal{F}_{t,1}\right]$$

$$\leq 3 \cdot \mathbb{E}\left[\sum_{i\in\mathcal{N}}\|r_{t+1}^i \phi_t\|^2 + \|\mu_t^i \phi_t\|^2 + \|\phi_t(\phi_{t+1}^\top - \phi_t^\top)\|^2 \|\omega_t^i\|^2 \Big| \mathcal{F}_{t,1}\right], \tag{5.8}$$

where the inequality follows from that by Assumption 4.4, $\mathbb{E}[\|\phi_t(\phi_{t+1}^\top - \phi_t^\top)\|^2 | \mathcal{F}_{t,1}]$ and $\mathbb{E}(\|\phi_t\|^2 | \mathcal{F}_{t,1})$ are both uniformly bounded for any $s_t \in \mathcal{S}$ and $a_t \in \mathcal{A}$. Moreover, by Assumption 4.2, we have $\mathbb{E}(|r_{t+1}^i|^2 | \mathcal{F}_{t,1}) = \mathbb{E}(|r_{t+1}^i|^2 | s_t, a_t)$ also uniformly bounded. Thus the RHS of (5.8) is bounded on the set $\{\sup_{\tau \leq t}\|z_\tau\| \leq M\}$ for any $M > 0$ as follows, i.e., there exists $K_1 < \infty$, such that

$$\mathbb{E}\left(\|y_{t+1}\|^2 \mathbb{I}_{\{\sup_{\tau\leq t}\|z_\tau\|\leq M\}} \Big| \mathcal{F}_{t,1}\right) \leq K_1 \cdot \left[1 + \sum_{i\in\mathcal{N}} \mathbb{E}\left(|r_{t+1}^i|^2 \Big| \mathcal{F}_{t,1}\right)\right]. \tag{5.9}$$

Let $\eta_t = \|\beta_{\omega,t}^{-1}\omega_{\perp,t}\|^2 \mathbb{I}_{\{\sup_{\tau\leq t}\|z_\tau\|\leq M\}}$ and note that $\mathbb{I}_{\{\sup_{\tau\leq t+1}\|z_\tau\|\leq M\}} \leq \mathbb{I}_{\{\sup_{\tau\leq t}\|z_\tau\|\leq M\}}$. Then by taking expectation over both sides of (5.7), we obtain that there exists $K_2 = K_1 \cdot [1 + \mathbb{E}(\sum_{i\in\mathcal{N}}|r_{t+1}^i|^2)] < \infty$ such that

$$\mathbb{E}(\eta_{t+1}) \leq \frac{\beta_{\omega,t}^2}{\beta_{\omega,t+1}^2} \cdot \rho \cdot \left[\mathbb{E}(\eta_t) + 2\sqrt{\mathbb{E}(\eta_t)} \cdot \sqrt{K_2} + K_2\right]. \tag{5.10}$$

Since $\lim_t \beta_{\omega,t}^2 \cdot \beta_{\omega,t+1}^{-2} = 1$ and $\rho < 1$, for any $\delta > 0$, there exists a large enough $t_0$ such that for any $t > t_0$, $\beta_{\omega,t}^2 \cdot \beta_{\omega,t+1}^{-2} \cdot \rho \leq 1 - \delta$. Hence, there exist positive constants $K_3$ and $b$ such that for any $t \geq t_0$,

$$\mathbb{E}(\eta_{t+1}) \leq (1 - \delta) \cdot \left[\mathbb{E}(\eta_t) + 2\sqrt{\mathbb{E}(\eta_t)} \cdot \sqrt{K_2} + K_2\right] \leq (1 - \delta/2) \cdot \mathbb{E}(\eta_t) + b \cdot \mathbb{I}_{\{\mathbb{E}(\eta_t)\leq K_3\}}.$$

By induction, we obtain that $\mathbb{E}(\eta_t) \leq (1 - \delta/2)^{t-t_0}\mathbb{E}(\eta_{t_0}) + 2b/\delta$. Hence, we have $\sup_t \mathbb{E}(\eta_t) < \infty$. In addition, since $\mathbb{I}_{\{\sup_t \|z_t\|\leq M\}} \leq \mathbb{I}_{\{\sup_{\tau\leq t}\|z_\tau\|\leq M\}}$, we further obtain

$$\sup_t \mathbb{E}\left(\|\beta_{\omega,t}^{-1}\omega_{\perp,t}\|^2 \mathbb{I}_{\{\sup_t\|z_t\|\leq M\}}\right) < \infty,$$

which concludes the proof. $\square$

Therefore, by Lemma 5.3, we obtain that for any $M > 0$, there exists a constant $K_4 < \infty$, such that for any $t \geq 0$, $\mathbb{E}(\|\omega_{\perp,t}\|^2) \leq K_4 \cdot \beta_{\omega,t}^2$ on the set $\{\sup_t\|z_t\| \leq M\}$. Since $\sum_t \beta_{\omega,t}^2 < \infty$ by Assumption 4.5, we have that $\sum_t \mathbb{E}(\|\omega_{\perp,t}\|^2 \mathbb{I}_{\{\sup_t\|z_t\|\leq M\}})$ is finite by Fubini's theorem. This shows that $\sum_t \|\omega_{\perp,t}\|^2 \mathbb{I}_{\{\sup_t\|z_t\|\leq M\}} < \infty$ a.s., which further yields $\lim_t \omega_{\perp,t}\mathbb{I}_{\{\sup_t\|z_t\|\leq M\}} = 0$ a.s. By Lemmas 5.1 and 5.2, $\{\sup_t\|z_t\| < \infty\}$ holds with probability 1. This shows that $\lim_t \omega_{\perp,t} = 0$ a.s., and thus concludes **Step 1**.

**Step 2.** We now proceed to show the convergence of the consensus vector $\mathbb{1} \otimes \langle \omega_t \rangle$. According to the update in (5.2) and definition (5.3), the iteration of $\langle \omega_t \rangle$ has the form

$$\langle \omega_{t+1} \rangle = \frac{1}{N}(\mathbb{1}^\top \otimes I)(C_t \otimes I)(\mathbb{1} \otimes \langle \omega_t \rangle + \omega_{t,\perp} + \beta_{\omega,t} y_{t+1}) = \langle \omega_t \rangle + \beta_{\omega,t} \langle (C_t \otimes I)(y_{t+1} + \beta_{\omega,t}^{-1} \omega_{\perp,t}) \rangle.$$

Hence, we write the updates for $\langle \omega_t \rangle$ and $\langle \mu_t \rangle$ as

$$\langle \mu_{t+1} \rangle = \langle \mu_t \rangle + \beta_{\omega,t} \cdot \mathbb{E}\big(\overline{r}_{t+1} - \langle \mu_t \rangle \big| \mathcal{F}_{t,1}\big) + \beta_{\omega,t} \cdot \xi_{t+1,1}, \tag{5.11}$$

$$\langle \omega_{t+1} \rangle = \langle \omega_t \rangle + \beta_{\omega,t} \cdot \mathbb{E}\big(\langle \delta_t \rangle \phi_t \big| \mathcal{F}_{t,1}\big) + \beta_{\omega,t} \cdot \xi_{t+1,2}, \tag{5.12}$$

where $\xi_{t+1,1} = \overline{r}_{t+1} - \mathbb{E}(\overline{r}_{t+1} | \mathcal{F}_{t,1})$ and $\xi_{t+1,2}$ is

$$\xi_{t+1,2} = \langle (C_t \otimes I)(y_{t+1} + \beta_{\omega,t}^{-1} \omega_{\perp,t}) \rangle - \mathbb{E}\big(\langle \delta_t \rangle \phi_t \big| \mathcal{F}_{t,1}\big).$$

Note that $\mathbb{E}(\overline{r}_{t+1} - \langle \mu_t \rangle | \mathcal{F}_{t,1})$ is Lipschitz continuous in $\langle \mu_t \rangle$. Recall that $\langle \delta_t \rangle$ has the form

$$\langle \delta_t \rangle = \frac{1}{N} \sum_{i \in \mathcal{N}} r_{t+1}^i - \mu_t^i + \phi_{t+1}^\top \omega_t^i - \phi_t^\top \omega_t^i = \overline{r}_{t+1} - \langle \mu_t \rangle + \phi_{t+1}^\top \langle \omega_t \rangle - \phi_t^\top \langle \omega_t \rangle.$$

Hence, $\mathbb{E}(\langle \delta_t \rangle \phi_t | \mathcal{F}_{t,1})$ is Lipschitz continuous in both $\langle \omega_t \rangle$ and $\langle \mu_t \rangle$, and thus the condition (a.1) in Assumption B.1 (See Appendix §B.1) is satisfied.

Note that $\xi_{t,1}$ is a martingale difference sequence and satisfies

$$\mathbb{E}\big(\|\xi_{t+1,1}\|^2 \big| \mathcal{F}_{t,1}\big) \leq K_5 \cdot \big(1 + \|\langle \omega_t \rangle\|^2 + \|\langle \mu_t \rangle\|^2\big), \tag{5.13}$$

for some $K_5 < \infty$, since $\overline{r}_{t+1}$ is uniformly bounded. In addition, the term $\xi_{t,2}$ is also a martingale difference sequence, since

$$\mathbb{E}\big[\langle (C_t \otimes I)(y_{t+1} + \beta_{\omega,t}^{-1} \omega_{\perp,t}) \rangle \big| \mathcal{F}_{t,1}\big] = \mathbb{E}\big[\langle (C_t \otimes I) y_{t+1} \rangle \big| \mathcal{F}_{t,1}\big] = \mathbb{E}\big(\langle y_{t+1} \rangle \big| \mathcal{F}_{t,1}\big) = \mathbb{E}\big(\langle \delta_t \rangle \phi_t \big| \mathcal{F}_{t,1}\big),$$

which results from the facts that $\langle \omega_{\perp,t} \rangle = 0$ and that $\mathbb{1}^\top \mathbb{E}(C_t) = \mathbb{1}^\top$. Moreover, we have

$$\mathbb{E}\big(\|\xi_{t+1,2}\|^2 \big| \mathcal{F}_{t,1}\big) \leq 2 \cdot \mathbb{E}\big(\big\|y_{t+1} + \beta_{\omega,t}^{-1} \omega_{\perp,t}\big\|_{G_t}^2 \big| \mathcal{F}_{t,1}\big) + 2 \cdot \big\|\mathbb{E}\big(\langle \delta_t \rangle \phi_t \big| \mathcal{F}_{t,1}\big)\big\|^2, \tag{5.14}$$

where $G_t = C_t^\top \mathbb{1}\mathbb{1}^\top C_t \otimes I \cdot N^{-2}$. Note that $G_t$ has bounded spectral norm since $C_t$ is a stochastic matrix. Thus the first term in (5.14) can be further bounded over the set $\{\sup_t \|z_t\| \leq M\}$, for any $M > 0$. Notably, there exist $K_6, K_7 < \infty$ such that

$$\mathbb{E}\big(\big\|y_{t+1} + \beta_{\omega,t}^{-1} \omega_{\perp,t}\big\|_{G_t}^2 \big| \mathcal{F}_{t,1}\big) \cdot \mathbb{I}_{\{\sup_t \|z_t\| \leq M\}} \leq K_6 \cdot \mathbb{E}\big(\|y_{t+1}\|^2 + \big\|\beta_{\omega,t}^{-1} \omega_{\perp,t}\big\|^2 \big| \mathcal{F}_{t,1}\big) \cdot \mathbb{I}_{\{\sup_t \|z_t\| \leq M\}} < K_7,$$

where the second inequality follows from (5.9) and Lemma 5.3. Moreover, the second term in (5.14) can be bounded by $\|\mathbb{E}(\langle \delta_t \rangle \phi_t | \mathcal{F}_{t,1})\|^2 \leq K_8 \cdot (1 + \|\langle \omega_t \rangle\|^2 + \|\langle \mu_t \rangle\|^2)$ with some $K_8 < \infty$, due to the boundedness of $r_{t+1}^i$ and $\phi_t$ (from Assumptions 4.2 and 4.4). Hence, for any $M > 0$, it follows that

$$\mathbb{E}\big(\|\xi_{t+1,1}\|^2 \big| \mathcal{F}_{t,1}\big) \leq K_9 \cdot \big(1 + \|\langle \omega_t \rangle\|^2 + \|\langle \mu_t \rangle\|^2\big), \tag{5.15}$$

over the set $\{\sup_t \|z_t\| \leq M\}$ for some $K_9 < \infty$. This verifies that on the set $\{\sup_t \|z_t\| \leq M\}$ for any $M > 0$, the condition (a.4) in Assumption B.1 is satisfied.

Now consider the following ODE that captures the asymptotic behavior of (5.11) and (5.12)

$$\langle \dot{z} \rangle = \begin{pmatrix} \langle \dot{\mu} \rangle \\ \langle \dot{\omega} \rangle \end{pmatrix} = \begin{pmatrix} -1 & 0 \\ -\Phi^\top D_\theta^{s,a} \mathbb{1} & \Phi^\top D_\theta^{s,a}(P^\theta - I)\Phi \end{pmatrix} \begin{pmatrix} \langle \mu \rangle \\ \langle \omega \rangle \end{pmatrix} + \begin{pmatrix} J(\theta) \\ \Phi^\top D_\theta^{s,a} \overline{R} \end{pmatrix}. \quad (5.16)$$

Recall that $D_\theta^{s,a} = \text{diag}[d_\theta(s) \cdot \pi_\theta(s,a), s \in \mathcal{S}, a \in \mathcal{A}]$. Let the RHS of the ODE (5.16) be $h(\langle z \rangle)$, then $h(\langle z \rangle)$ is Lipschitz continuous in $\langle z \rangle$, which satisfies the condition (a.1) in Assumption B.1. By the Perron-Frobenius theorem and Assumption 2.2, the stochastic matrix $P^\theta$ has a simple eigenvalue of 1, and its remaining eigenvalues have real parts less than 1. Hence, $(P^\theta - I)$ has all eigenvalues with negative real parts but one zero, so does the matrix $\Phi^\top D_\theta^{s,a}(P^\theta - I)\Phi$, since $\Phi$ is full column rank by Assumption 4.4. The simple eigenvalue of zero has eigen-vector $v$ that satisfies $\Phi v = \alpha \mathbb{1}$ for some $\alpha \neq 0$, since $\alpha \mathbb{1}$ lies in the eigen-space of $D_\theta^{s,a}(P^\theta - I)$ associated with zero. By Assumption 4.4, however, this will not happen with any choice of $\Phi$ since $\Phi v \neq \alpha \mathbb{1}$ for any $v \in \mathbb{R}^K$. Hence, the ODE (5.16) is globally asymptotically stable and has its equilibrium satisfying

$$-\langle \mu \rangle = J(\theta), \qquad \Phi^\top D_\theta^{s,a}\left[\overline{R} - \langle \mu \rangle \mathbb{1} + P^\theta \Phi \langle \omega \rangle - \Phi \langle \omega \rangle\right] = 0. \quad (5.17)$$

Note that the corresponding solution for $\langle \mu \rangle$ at equilibrium is $J(\theta)$, whereas the solution for $\langle \omega \rangle$ has the form $\omega_\theta + \alpha v$ with any $\alpha \in \mathbb{R}$ and $v \in \mathbb{R}^K$ such that $\Phi v = \mathbb{1}$. While by Assumption 4.4, $\Phi v \neq \mathbb{1}$, thus the term $\omega_\theta$ is unique, and it follows that $\Phi^\top D_\theta^{s,a}[T_\theta^Q(\Phi \omega_\theta) - \Phi \omega_\theta] = 0$ with $T_\theta^Q$ as defined in (4.1).

Recall from Lemmas 5.1 and 5.2 that $\{z_t\}$ is bounded a.s., so is the sequence $\{\langle z_t \rangle\}$. Hence all conditions for Theorem B.2 to hold are satisfied. For the concatenated vector $\langle z_t \rangle = (\langle \mu_t \rangle, \langle \omega_t \rangle^\top)^\top$, we thus have $\lim_t \langle \mu_t \rangle = J(\theta)$ and $\lim_t \langle \omega_t \rangle = \omega_\theta$ over the set $\{\sup_t \|z_t\| \leq M\}$ for any $M > 0$. By Lemmas 5.1 and 5.2, this holds with probability 1, which concludes **Step 2**. Combined with **Step 1**, we arrive at the conclusion that $\lim_t \omega_t^i = \omega_\theta$ for any $i \in \mathcal{N}$, which completes the proof for Theorem 4.6. □

## 5.2 Proof of Theorem 4.7

Let $\mathcal{F}_{t,2} = \sigma(\theta_\tau, \tau \leq t)$ be the $\sigma$-field generated by $\{\theta_\tau, \tau \leq t\}$. In addition, we define

$$\zeta_{t+1,1}^i = A_t^i \cdot \psi_t^i - \mathbb{E}_{s_t \sim d_{\theta_t}, a_t \sim \pi_{\theta_t}}\left(A_t^i \cdot \psi_t^i \mid \mathcal{F}_{t,2}\right), \qquad \zeta_{t+1,2}^i = \mathbb{E}_{s_t \sim d_{\theta_t}, a_t \sim \pi_{\theta_t}}\left[\left(A_t^i - A_{t,\theta_t}^i\right) \cdot \psi_t^i \mid \mathcal{F}_{t,2}\right],$$

where $A_{t,\theta_t}^i$ is as defined in (4.4) with $\theta = \theta_t$. Then the actor update in (3.9) with a local projection becomes

$$\theta_{t+1}^i = \Gamma^i\left[\theta_t^i + \beta_{\theta,t}\mathbb{E}_{s_t \sim d_{\theta_t}, a_t \sim \pi_{\theta_t}}\left(A_{t,\theta_t}^i \cdot \psi_t^i \mid \mathcal{F}_{t,2}\right) + \beta_{\theta,t}\zeta_{t+1,1}^i + \beta_{\theta,t}\zeta_{t+1,2}^i\right]. \quad (5.18)$$

Note that $\zeta_{t+1,2}^i = o(1)$ since the critic converges, i.e., $A_t^i \to A_{t,\theta_t}^i$, at the faster time scale. Moreover, letting $M_t^i = \sum_{\tau=0}^t \beta_{\theta,\tau}\zeta_{\tau+1,1}^i$, $\{M_t^i\}$ is a martingale sequence. Since the sequences $\{\omega_t^i\}$, $\{\psi_t^i\}$, and $\{\phi_t\}$ are all bounded, the sequence $\{\zeta_{t,1}^i\}$ is also bounded. Hence, by Assumption 4.5, we have

$$\sum_t \mathbb{E}\left(\|M_{t+1}^i - M_t^i\|^2 \mid \mathcal{F}_{t,2}\right) = \sum_{t \geq 1} \|\beta_{\theta,t}\zeta_{t+1,1}^i\|^2 < \infty \quad \text{a.s.}$$

By the martingale convergence theorem (Proposition VII-2-3(c) on page 149 of Neveu (1975)), the martingale sequence $\{M_t^i\}$ converges a.s. Thus, for any $\epsilon > 0$, we have

$$\lim_t \mathbb{P}\left(\sup_{n \geq t} \left\|\sum_{\tau=t}^n \beta_{\theta,\tau} \zeta_{\tau,1}^i\right\| \geq \epsilon\right) = 0.$$

In addition, let

$$g^i(\theta_t) = \mathbb{E}_{s_t \sim d_{\theta_t}, a_t \sim \pi_{\theta_t}}\left(A_{t,\theta_t}^i \cdot \psi_t^i \,\big|\, \mathcal{F}_{t,2}\right) = \sum_{s_t \in \mathcal{S}, a_t \in \mathcal{A}} d_{\theta_t}(s_t) \cdot \pi_{\theta_t}(s_t, a_t) \cdot A_{t,\theta_t}^i \cdot \psi_{t,\theta_t}^i,$$

then we show that $g^i(\theta_t)$ is continuous in $\theta_t^i$ as follows. First, $\psi_{t,\theta_t}^i$ is continuous by Assumption 2.2. Also, the term $d_{\theta_t}(s_t) \cdot \pi_{\theta_t}(s_t, a_t)$ is continuous in $\theta_t^i$ since it is the stationary distribution and thus is the solution to $d_{\theta_t}(s) \cdot \pi_{\theta_t}(s, a) = \sum_{s' \in \mathcal{S}, a' \in \mathcal{A}} P^{\theta_t}(s', a' | s, a) \cdot d_{\theta_t}(s') \cdot \pi_{\theta_t}(s', a')$ and $\sum_{s \in \mathcal{S}, a \in \mathcal{A}} d_{\theta_t}(s) \cdot \pi_{\theta_t}(s, a) = 1$, where $P^{\theta_t}(s', a' | s, a) = P(s' | s, a) \cdot \pi_{\theta_t}(s', a')$. The unique solution to this set of linear equations can be verified to be continuous in $\theta_t$, noting that $\pi_{\theta_t}(s, a) > 0$ by Assumption 2.2. In addition, $A_{t,\theta_t}^i$ is continuous in $\theta_t^i$ since $\omega_{\theta_t}$ is the unique solution to the linear equation $\Phi^\top D_\theta^{s,a}[T_\theta^Q(\Phi\omega_\theta) - \Phi\omega_\theta] = 0$ and can also be verified to be continuous in $\theta_t$. Therefore, by Kushner-Clark lemma (Kushner and Clark, 1978, page 191-196) (see also Theorem B.5 in Appendix §B.2), the update in (5.18) converges a.s. to the set of asymptotically stable equilibria of the ODE (4.6) for each $i \in \mathcal{N}$, which concludes the proof. □

The proof for the convergence of Algorithm 2 is similar to the proofs in 5.1 and 5.2. To avoid duplication, we leave out some of the details in the proofs to follow.

## 5.3 Proof of Theorem 4.10

Let $z_t^i = [\mu_t^i, (\lambda_t^i)^\top, (v_t^i)^\top]^\top \in \mathbb{R}^{1+M+L}$. We first have the following lemma on the stability of the updates of $\{z_t^i\}$, as in Lemma 5.1, the proof of which is provided in Appendix §A.

**Lemma 5.4.** *Under Assumptions 2.2, 4.2, 4.3, 4.8, and 4.9, the sequence $\{z_t^i\}$ generated from (3.12), (3.13), and (3.16) satisfies $\sup_t \|z_t^i\| < \infty$ a.s., for any $i \in \mathcal{N}$.*

Note that $\widetilde{\delta}_t^i \cdot \psi_t^i$ is bounded by Assumptions 2.2, 4.8, and Lemma 5.4. Thus the actor step (3.17) can be viewed to track ODE $\dot{\theta}^i = 0$ when analyzing the faster time scale update. Thus by the same argument as in §5.1, we fix the value of $\theta_t$ as a constant $\theta$. For notational convenience, let $v_t = [(v_t^1)^\top, \cdots, (v_t^N)^\top]^\top$, $\delta_t = [(\delta_t^1)^\top, \cdots, (\delta_t^N)^\top]^\top$, and $z_t = [(z_t^1)^\top, \cdots, (z_t^N)^\top]^\top$. We also use $\{\mathcal{F}_{t,1}\}$ to denote the filtration with $\mathcal{F}_{t,1} = \sigma(r_\tau, z_\tau, s_\tau, C_{\tau-1}, \tau \leq t)$, an increasing $\sigma$-algebra. Then the updates of $z_t$ in (3.12), (3.13), and (3.16) have the following compact form

$$z_{t+1} = (C_t \otimes I)(z_t + \beta_{v,t} \cdot y_{t+1}), \tag{5.19}$$

where $y_t = [(y_t^1)^\top, \cdots, (y_t^N)^\top]^\top \in \mathbb{R}^{(1+M+L)N}$. Here we denote $[r_{t+1}^i - \mu_t^i, (r_{t+1}^i - f_t^\top \lambda_t^i) f_t^\top, \delta_t^i \varphi_t^\top]^\top$ by $y_{t+1}^i$. Recall that $f_t = f(s_t, a_t)$ and $\varphi_t = \varphi(s_t)$. With the same definitions for $\langle \cdot \rangle$, $\mathcal{J}$, and $\mathcal{J}_\perp$, we can also separate the iteration of $z_t$ as the sum of the consensus vector and the disagreement vector, i.e., $z_t = z_{\perp,t} + \mathbb{1} \otimes \langle z_t \rangle$, with $z_{\perp,t} = \mathcal{J}_\perp z_t$. Then the proof proceeds again in two steps as follows.

**Step 1.** We first establish that $\lim_t z_{\perp,t} = 0$ a.s. By Lemma 5.4, it suffices to show that for any $M \in \mathbb{Z}^+$, $\lim_t z_{\perp,t} \mathbb{I}_{\{\sup_t \|z_t\| \leq M\}} = 0$. We first establish the boundedness of $\mathbb{E}(\|\beta_{v,t}^{-1} z_{\perp,t}\|^2)$ over the set $\{\sup_t \|z_t\| \leq M\}$, for any $M > 0$.

**Lemma 5.5.** Under Assumptions 4.2, 4.3, 4.8, and 4.9, for any $M > 0$, we have

$$\sup_t \mathbb{E}\left(\|\beta_{v,t}^{-1} z_{\perp,t}\|^2 \mathbb{I}_{\{\sup_t \|z_t\| \leq M\}}\right) < \infty.$$

*Proof.* Following the derivation of (5.6), we obtain the iteration of $z_{\perp,t}$ as

$$z_{\perp,t+1} = [(I - \mathbb{1}\mathbb{1}^\top/N) \otimes I](C_t \otimes I)(z_{\perp,t} + \beta_{v,t} y_{t+1}) = [(I - \mathbb{1}\mathbb{1}^\top/N) C_t \otimes I](z_{\perp,t} + \beta_{v,t} y_{t+1}). \quad (5.20)$$

Thus, similar to the derivation of (5.7), we obtain

$$\mathbb{E}\left(\|\beta_{v,t+1}^{-1} z_{\perp,t+1}\|^2 \big| \mathcal{F}_{t,1}\right) \leq \rho \cdot \frac{\beta_{v,t}^2}{\beta_{v,t+1}^2} \cdot \Big\{ \|\beta_{v,t}^{-1} z_{\perp,t}\|^2 + 2\|\beta_{v,t}^{-1} z_{\perp,t}\|$$

$$\cdot \left[\mathbb{E}\left(\|y_{t+1}\|^2 \big| \mathcal{F}_{t,1}\right)\right]^{\frac{1}{2}} + \mathbb{E}\left(\|y_{t+1}\|^2 \big| \mathcal{F}_{t,1}\right)\Big\}, \quad (5.21)$$

where $\rho$ represents the spectral norm of $\mathbb{E}[C_t^\top (I - \mathbb{1}\mathbb{1}^\top/N) C_t]$ and $\rho \in [0, 1)$. Then we have

$$\mathbb{E}\left(\|y_{t+1}\|^2 \big| \mathcal{F}_{t,1}\right) = \mathbb{E}\left[\sum_{i \in \mathcal{N}} |r_{t+1}^i - \mu_t^i|^2 + \|(r_{t+1}^i - f_t^\top \lambda_t^i) f_t\|^2 + \|\delta_t^i \varphi_t\|^2 \Big| \mathcal{F}_{t,1}\right]. \quad (5.22)$$

By Assumption 4.8, we have $\mathbb{E}[\|\varphi_t(\varphi_{t+1}^\top - \varphi_t^\top)\|^2 | \mathcal{F}_{t,1}]$ and $\mathbb{E}(\|\varphi_t\|^2 | \mathcal{F}_{t,1})$ uniformly bounded for any $s_t \in \mathcal{S}$. Moreover, by Assumption 4.2, we have $\mathbb{E}(|r_{t+1}^i|^2 | \mathcal{F}_{t,1}) = \mathbb{E}(|r_{t+1}^i|^2 | s_t, a_t)$ also uniformly bounded for any $s_t \in \mathcal{S}, a_t \in \mathcal{A}$. Thus for any $M > 0$, there exist $K_1, K_2 < \infty$[4] such that (5.22) is further bounded as

$$\mathbb{E}\left(\|y_{t+1}\|^2 \mathbb{I}_{\{\sup_{\tau \leq t} \|z_\tau\| \leq M\}} \big| \mathcal{F}_{t,1}\right) \leq K_1 \cdot \left[1 + \sum_{i \in \mathcal{N}} \mathbb{E}\left(|r_{t+1}^i|^2 \big| \mathcal{F}_{t,1}\right)\right] < K_2. \quad (5.23)$$

Let $\eta_t = \|\beta_{v,t}^{-1} z_{\perp,t}\|^2 \mathbb{I}_{\{\sup_{\tau \leq t} \|z_\tau\| \leq M\}}$. By taking expectation on both sides of (5.21), we obtain

$$\mathbb{E}(\eta_{t+1}) \leq \frac{\beta_{v,t}^2}{\beta_{v,t+1}^2} \cdot \rho \cdot \left[\mathbb{E}(\eta_t) + 2\sqrt{\mathbb{E}(\eta_t)} \cdot \sqrt{K_2} + K_2\right].$$

Following the same argument as in the proof of Lemma 5.3, we obtain $\sup_t \mathbb{E}(\eta_t) < \infty$ and thus

$$\sup_t \mathbb{E}\left(\|\beta_{v,t}^{-1} z_{\perp,t}\|^2 \mathbb{I}_{\{\sup_t \|z_t\| \leq M\}}\right) < \infty,$$

which concludes the proof. □

Therefore, by Lemma 5.5 and Assumption 4.9, we arrive at $\lim_t z_{\perp,t} \mathbb{I}_{\{\sup_t \|z_t\| \leq M\}} = 0$ a.s. for any $M > 0$. By Lemma 5.4, $\{\sup_t \|z_t\| < \infty\}$ holds with probability 1. This shows that $\lim_t z_{\perp,t} = 0$ a.s., and thus concludes **Step 1**.

---

[4] We note that $K_1$ and $K_2$ here are absolute constant values, with slight abuse of notation, we use the same notation as in the proof of Theorem 4.6. The same abuse applies to the other constants with notation $K_j$, for any $j \in \mathbb{N}$.

**Step 2.** We now proceed to show the convergence of the consensus vector $\mathbb{1} \otimes \langle z_t \rangle$. The iteration of $\langle z_t \rangle$ has the form

$$\langle z_{t+1} \rangle = \frac{1}{N}(\mathbb{1}^\top \otimes \mathrm{I})(\mathrm{C}_t \otimes \mathrm{I})(\mathbb{1} \otimes \langle z_t \rangle + z_{t,\perp} + \beta_{v,t} y_{t+1}) = \langle z_t \rangle + \beta_{v,t} \langle (\mathrm{C}_t \otimes \mathrm{I})(y_{t+1} + \beta_{v,t}^{-1} z_{\perp,t}) \rangle.$$

Hence, the update for $\langle z_t \rangle$ becomes

$$\langle z_{t+1} \rangle = \langle z_t \rangle + \beta_{v,t} \cdot \mathbb{E}(\langle y_{t+1} \rangle | \mathcal{F}_{t,1}) + \beta_{v,t} \cdot \xi_{t+1}, \tag{5.24}$$

where $\xi_{t+1}$ and $\langle y_{t+1} \rangle$ have the form

$$\xi_{t+1} = \langle (\mathrm{C}_t \otimes \mathrm{I})(y_{t+1} + \beta_{\omega,t}^{-1} \omega_{\perp,t}) \rangle - \mathbb{E}(\langle y_{t+1} \rangle | \mathcal{F}_{t,1}), \qquad \langle y_{t+1} \rangle = [\overline{r}_{t+1} - \langle \mu_t \rangle, (\overline{r}_{t+1} - f_t^\top \langle \lambda_t \rangle) f_t^\top, \langle \delta_t \rangle \varphi_t^\top]^\top,$$

respectively. Recall that $\langle \delta_t \rangle = \overline{r}_{t+1} - \langle \mu_t \rangle + \varphi_{t+1}^\top \langle v_t \rangle - \varphi_t^\top \langle v_t \rangle$. Note that $\mathbb{E}(\langle y_{t+1} \rangle | \mathcal{F}_{t,1})$ is Lipschitz continuous in $\langle z_t \rangle = (\langle \mu_t \rangle, \langle \lambda_t \rangle^\top, \langle v_t \rangle^\top)^\top$, and thus the condition (a.1) in Assumption B.1 is satisfied. Moreover, one can verify that the term $\xi_t$ is a martingale difference sequence. The conditional second moment of $\xi_t$ can be bounded as

$$\mathbb{E}\big(\|\xi_{t+1}\|^2 \big| \mathcal{F}_{t,1}\big) \leq 2 \cdot \mathbb{E}\big(\big\|y_{t+1} + \beta_{v,t}^{-1} z_{\perp,t}\big\|_{\mathrm{G}_t}^2 \big| \mathcal{F}_{t,1}\big) + 2 \cdot \big\|\mathbb{E}(\langle y_{t+1} \rangle | \mathcal{F}_{t,1})\big\|^2, \tag{5.25}$$

where $\mathrm{G}_t = \mathrm{C}_t^\top \mathbb{1}\mathbb{1}^\top \mathrm{C}_t \otimes \mathrm{I} \cdot N^{-2}$ has bounded spectral norm. Thus the first term in (5.25) is bounded over the set $\{\sup_t \|z_t\| \leq M\}$, for any $M > 0$, i.e., there exist $K_3, K_4 < \infty$ such that

$$\mathbb{E}\big(\big\|y_{t+1} + \beta_{v,t}^{-1} z_{\perp,t}\big\|_{\mathrm{G}_t}^2 \big| \mathcal{F}_{t,1}\big) \cdot \mathbb{I}_{\{\sup_t \|z_t\| \leq M\}} \leq K_3 \cdot \mathbb{E}\big(\|y_{t+1}\|^2 + \big\|\beta_{v,t}^{-1} z_{\perp,t}\big\|^2 \big| \mathcal{F}_{t,1}\big) \cdot \mathbb{I}_{\{\sup_t \|z_t\| \leq M\}} < K_4, \tag{5.26}$$

where the second inequality follows from (5.23) and Lemma 5.5. Moreover, the second term in (5.25) can be bounded by

$$\big\|\mathbb{E}(\langle y_{t+1} \rangle | \mathcal{F}_{t,1})\big\|^2 \leq \mathbb{E}\big(\|\langle y_{t+1} \rangle\|^2 \big| \mathcal{F}_{t,1}\big) \leq K_5 \cdot \big(1 + \|\langle \mu_t \rangle\|^2 + \|\langle \lambda_t \rangle\|^2 + \|\langle v_t \rangle\|^2\big) = K_5 \cdot \big(1 + \|\langle z_t \rangle\|^2\big)$$

with some $K_5 < \infty$ due to the boundedness of $r_{t+1}^i$, $f_t$, and $\varphi_t$. Hence, for any $M > 0$, it follows that

$$\mathbb{E}\big(\|\xi_{t+1}\|^2 \big| \mathcal{F}_{t,1}\big) \leq K_6 \cdot \big(1 + \|\langle z_t \rangle\|^2\big), \tag{5.27}$$

over the set $\{\sup_t \|z_t\| \leq M\}$ for some $K_6 < \infty$. This verifies the condition (a.4) in Assumption B.1.

Then the ODE associated with (5.24) has the form

$$\langle \dot{z} \rangle = \begin{pmatrix} \langle \dot{\mu} \rangle \\ \langle \dot{\lambda} \rangle \\ \langle \dot{v} \rangle \end{pmatrix} = \begin{pmatrix} -1 & 0 & 0 \\ 0 & -\mathrm{F}^\top \mathrm{D}_\theta^{s,a} \mathrm{F} & 0 \\ -\Phi^\top \mathrm{D}_\theta^s \mathbb{1} & 0 & \Phi^\top \mathrm{D}_\theta^s (P^\theta - \mathrm{I}) \Phi \end{pmatrix} \begin{pmatrix} \langle \mu \rangle \\ \langle \lambda \rangle \\ \langle v \rangle \end{pmatrix} + \begin{pmatrix} J(\theta) \\ \mathrm{F}^\top \mathrm{D}_\theta^{s,a} \overline{R} \\ \Phi^\top \mathrm{D}_\theta^s \overline{R}_\theta \end{pmatrix}. \tag{5.28}$$

Letting the RHS of the ODE (5.28) be $h(\langle z \rangle)$, we have $h(\langle z \rangle)$ Lipschitz continuous in $\langle z \rangle$. Similar to the proof in **Step 2** of §5.1, one can verify that the ODE has a unique globally asymptotically stable equilibrium $[J(\theta), \lambda_\theta^\top, v_\theta^\top]^\top$, by Assumption 4.8 on the feature matrices F and $\Phi$. Here $\lambda_\theta$ and $v_\theta$ are the unique solutions to $\mathrm{F}^\top \mathrm{D}_\theta^{s,a}(\overline{R} - \mathrm{F}\lambda_\theta) = 0$ and $\Phi^\top \mathrm{D}_\theta^s \big[T_\theta^V(\Phi v_\theta) - \Phi v_\theta\big] = 0$, respectively. Recall the operator $T_\theta^V$ defined in (4.7). Moreover, the sequence $\{z_t\}$ is bounded almost surely by Assumption 5.4. Hence all conditions for Theorem B.2 to hold are satisfied. We thus have $\lim_t \langle \mu_t \rangle = J(\theta)$, $\lim_t \langle \lambda_t \rangle = \lambda_\theta$, and $\lim_t \langle v_t \rangle = v_\theta$ over the set $\{\sup_t \|z_t\| \leq M\}$ for any $M > 0$. By Lemma 5.4 and the results from **Step 1**, we obtain that $\lim_t \mu_t^i = J(\theta)$, $\lim_t \lambda_t^i = \lambda_\theta$, and $\lim_t v_t^i = v_\theta$ for any $i \in \mathcal{N}$ a.s., which completes the proof. □

## 5.4 Proof of Theorem 4.11

Let $\mathcal{F}_{t,2} = \sigma(\theta_\tau, \tau \leq t)$ be the $\sigma$-field generated by $\theta_\tau, \tau \leq t$. Let

$$\zeta^i_{t+1,1} = \widetilde{\delta}^i_t \cdot \psi^i_t - \mathbb{E}_{s_t \sim d_{\theta_t}, a_t \sim \pi_{\theta_t}}\big(\widetilde{\delta}^i_t \cdot \psi^i_t \big| \mathcal{F}_{t,2}\big), \qquad \zeta^i_{t+1,2} = \mathbb{E}_{s_t \sim d_{\theta_t}, a_t \sim \pi_{\theta_t}}\big[\big(\widetilde{\delta}^i_t - \widetilde{\delta}^i_{t,\theta_t}\big) \cdot \psi^i_t \big| \mathcal{F}_{t,2}\big],$$

where $\widetilde{\delta}^i_{t,\theta_t}$ is as defined in (4.9) with $\theta = \theta_t$. Then the actor update in (3.17) with a local projection becomes

$$\theta^i_{t+1} = \Gamma^i\bigg[\theta^i_t + \beta_{\theta,t}\mathbb{E}_{s_t \sim d_{\theta_t}, a_t \sim \pi_{\theta_t}}\big(\widetilde{\delta}^i_t \cdot \psi^i_t \big| \mathcal{F}_{t,2}\big) + \beta_{\theta,t}\zeta^i_{t+1,1} + \beta_{\theta,t}\zeta^i_{t+1,2}\bigg]. \tag{5.29}$$

Note that $\zeta^i_{t+1,2} = o(1)$ since the critic converges, i.e., $\widetilde{\delta}^i_t \to \widetilde{\delta}^i_{t,\theta_t}$, at the faster time scale. Moreover, letting $M^i_t = \sum_{\tau=0}^{t} \beta_{\theta,\tau} \zeta^i_{\tau+1,1}$, we have $\{M^i_t\}$ a martingale sequence. Note that the sequences $\{z^i_t\}, \{\psi^i_t\}$, and $\{\phi_t\}$ are all bounded, and so is the sequence $\{\zeta^i_{t,1}\}$. Hence, we have $\sum_t \mathbb{E}(\|M^i_{t+1} - M^i_t\|^2 | \mathcal{F}_{t,2}) < \infty$ a.s., and further obtain that the martingale sequence $\{M^i_t\}$ converges a.s. (Neveu, 1975, page 149)). Thus the condition (a.4) in Assumption B.4 is satisfied. (See Appendix §B.1 for details.) One can also verify that $\mathbb{E}_{s_t \sim d_{\theta_t}, a_t \sim \pi_{\theta_t}}(\widetilde{\delta}^i_t \cdot \psi^i_t | \mathcal{F}_{t,2})$ is continuous in $\theta^i_t$. Therefore, we can apply the Kushner-Clark lemma (Theorem B.5 in Appendix §B.2) to show that the update in (5.29) converges a.s. to the set of asymptotically stable equilibria of the ODE (4.10), for each $i \in \mathcal{N}$, which concludes the proof. □

## 6 Numerical Results

In this section, we evaluate the proposed fully decentralized AC algorithms through numerical simulations with both linear and nonlinear function approximation.

### 6.1 Linear Function Approximation

We first consider the setting where linear function approximation is adopted. Consider in total $N = 20$ agents, each has a binary-valued action space, i.e., $\mathcal{A}^i = \{0,1\}$, for all $i \in \mathcal{N}$. Thus the cardinality of the set of actions $\mathcal{A}$ is $2^{20}$. In addition, there are in total $|\mathcal{S}| = 20$ states. The following ways of selecting the model and algorithm parameters, including transition probabilities, rewards, and features, follow from those in Dann et al. (2014). The elements in the transition probability matrix $P$ are uniformly sampled from the interval $[0,1]$ and normalized to be stochastic. We also add a small constant $10^{-5}$ onto each element in the matrix to ensure ergodicity of the MDP such that Assumption 2.2 is satisfied. For each agent $i$ and each state-action pair $(s,a)$, the mean reward $R^i(s,a)$ is sampled uniformly from $[0,4]$, which varies among agents. The instantaneous rewards $r^i_t$ are sampled from the uniform distribution $[R^i(s,a) - 0.5, R^i(s,a) + 0.5]$. The policy $\pi^i_{\theta^i}(s, a^i)$ is parameterized following the Boltzman policies, i.e.,

$$\pi^i_{\theta^i}(s, a^i) = \frac{\exp\big(q^\top_{s,a^i}\theta^i\big)}{\sum_{b^i \in \mathcal{A}^i} \exp\big(q^\top_{s,b^i}\theta^i\big)}$$

where $q_{s,b^i} \in \mathbb{R}^{m_i}$ is the feature vector with the same dimension as $\theta^i$, for any $s \in \mathcal{S}$ and $i \in \mathcal{N}$. Here we set $m_1 = m_2 = \cdots = m_N = 5$. The elements of $q_{s,b^i}$ are also uniformly sampled from $[0,1]$. In particular, the gradient of the score function thus has the form

$$\nabla_{\theta^i} \log \pi^i_{\theta^i}(s, a^i) = q_{s,a^i} - \sum_{b^i \in \mathcal{A}^i} \pi^i_{\theta^i}(s, a^i) q_{s,b^i}.$$

The feature vectors $\phi \in \mathbb{R}^K$ for the action-value function $Q(\cdot,\cdot;\omega)$ in Algorithm 1, $\varphi \in \mathbb{R}^L$ for the value function $V(\cdot;v)$ and $f \in \mathbb{R}^M$ for the globally averaged reward function $\overline{R}(\cdot,\cdot;\lambda)$ in Algorithm 2, are all uniformly sampled from $[0,1]$, of dimensions $K = 10 \ll |\mathcal{S}| \cdot |\mathcal{A}|$, $L = 5 < |\mathcal{S}|$, and $M = 10 \ll |\mathcal{S}| \cdot |\mathcal{A}|$. Moreover, the selected feature matrices $\Phi$, $\varPhi$, and $F$ are all ensured to have full column rank as required in Assumptions 4.4 and 4.8.

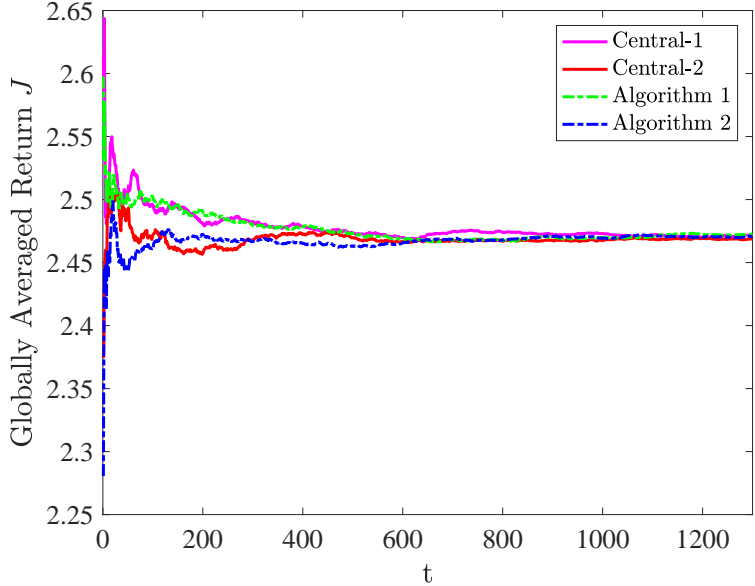

Figure 1: The convergence of globally averaged returns, when linear function approximation is used. We plot the returns achieved by both Algorithm 1 and Algorithm 2, along with their centralized counterparts Central-1 and Central-2.

The consensus weight matrix $C_t$ is chosen independent and identically distributed along time $t$ by normalizing the absolute Laplacian matrix of a random graph $\mathcal{G}_t$ over agents $\mathcal{N}$ to be doubly stochastic. The graph $\mathcal{G}_t$ is generated by randomly placing communication links among agents such that the connectivity ratio[5] is $4/N$. The stepsizes are selected as $\beta_{\omega,t} = \beta_{v,t} = 1/t^{0.65}$ and $\beta_{\theta,t} = 1/t^{0.85}$, which satisfy Assumptions 4.5 and 4.9.

The performances of the fully decentralized algorithms are compared with those of the centralized algorithms in which the rewards $r^i_t$ of all agents are available at a centralized controller and the global policy $\pi_\theta$ is also updated there. These centralized version algorithms thus reduce to single-agent AC algorithms with linear function approximation. We use two centralized AC algorithms based on action-value and state-value function approximation, to compare with Algorithm 1 and Algorithm 2, respectively. In particular, the former one, referred to as Central-1, follows the single-agent AC updates in (2.3). The later one, referred to as Central-2, follows the updates of

---

[5]Note that the connectivity ratio is defined as the ratio between the total degree of the graph and the degree of the complete graph, i.e., $2E/[N(N-1)]$, where $E$ is the number of edges.

Algorithm 1 in Bhatnagar et al. (2009) and is based on state-value TD-error as Algorithm 2 here. In particular, it has the following critic step

$$\begin{cases} \mu_{t+1} = (1-\beta_{v,t})\cdot \mu_t + \beta_{v,t}\cdot \bar{r}_{t+1}, \\ \delta_t = \bar{r}_{t+1} - \mu_t + V_{t+1}(v_t) - V_t(v_t), \\ v_{t+1} = v_t + \beta_{v,t}\cdot \delta_t \cdot \nabla_v V_t(v_t), \end{cases} \quad (6.1)$$

where we recall that $V_t(v) = V(s_t; v)$ for any $v \in \mathbb{R}^L$ and $\beta_{v,t} > 0$ is the stepsize satisfying Assumption 4.5. Since the rewards of all agents $\{r_t^i\}_{i \in \mathcal{N}}$ are available to the controller, no estimation for the globally averaged reward function $\bar{R}$ is needed in the update. Thus, the global state-value TD-error $\delta_t$ can be computed immediately and used in the actor step as

$$\theta_{t+1}^i = \theta_t^i + \beta_{\theta,t}\cdot \delta_t \cdot \psi_t^i, \quad \forall i \in \mathcal{N}, \quad (6.2)$$

where $\psi_t^i$ is as defined in (3.10).

As shown in Figure 1, both decentralized algorithms converge to the globally long-term averaged return as achieved by the two centralized algorithms. Moreover, it is corroborated in Figure 2 that for each agent, the approximation of action-value function in Algorithm 1 and state-value function in Algorithm 2 reach consensus at a much faster rate than the AC algorithm converges. It is also observed that the decentralized algorithms converge to the stationary relative value functions slower than the centralized counterparts, possibly due to the delay of information diffusion across the network.

In addition, in terms of policy, both Algorithms 1 and 2 converge to similar policies as their central counterparts. Figure 3 illustrates the resulting policies by all algorithms, i.e., the probability distribution $\pi_{\theta^i}^i(s, a^i)$ for a randomly selected state $s$ for any $i \in \mathcal{N}$. Note that there are in total $2 \times 20$ probability values under our setting with binary-valued actions and $|\mathcal{S}| = 20$ states. It is verified that the joint policy obtained by agents using only local information is almost as good as the policy obtained by the centralized controller with full system information.

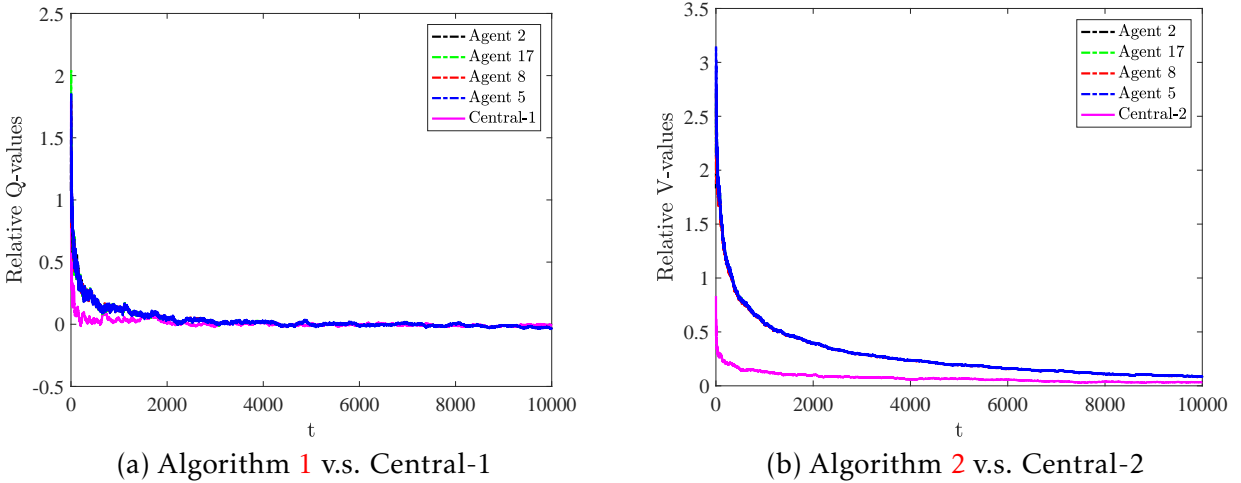

(a) Algorithm 1 v.s. Central-1

(b) Algorithm 2 v.s. Central-2

Figure 2: The convergence of relative value functions at four randomly selected agents, when linear function approximation is used. We randomly select the agents 2, 5, 8, and 17. In (a), we plot the convergence curve of the relative action-value at a randomly selected state-action pair, obtained from Central-1 and Algorithm 1. In (b), we plot the convergence curve of the relative state-value at a randomly selected state, obtained from Central-2 and Algorithm 2.

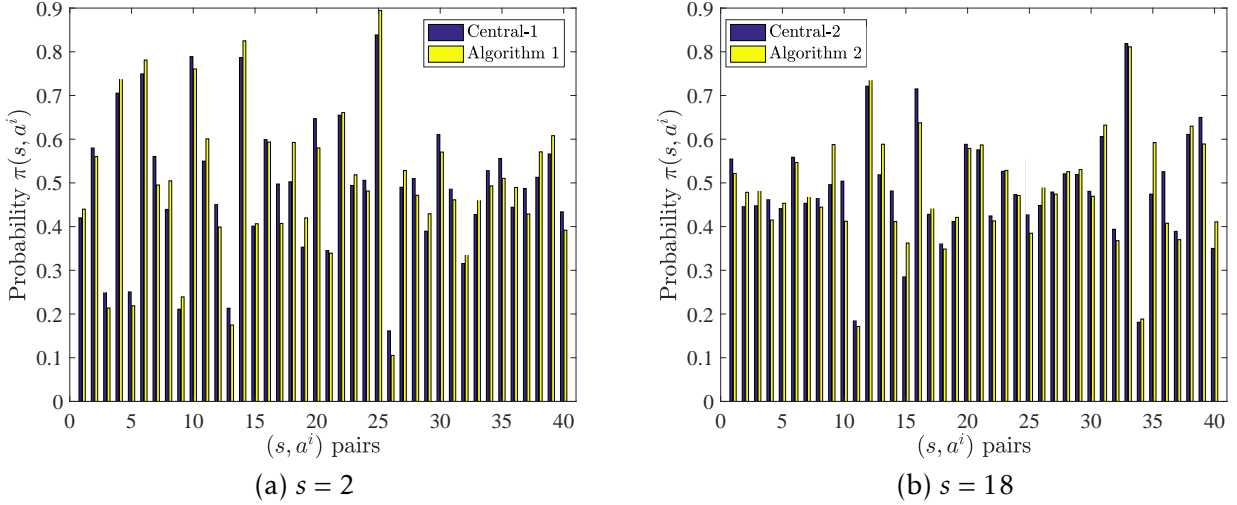

Figure 3: The policies, i.e., the probability distribution $\pi_\theta(s,a)$ at a randomly selected state $s$. In (a), we plot the policy obtained from Central-1 and Algorithm 1 at state 2. In (b), we plot the policy obtained from Central-2 and Algorithm 2 at state 18.

## 6.2 Nonlinear Function Approximation

We also empirically evaluate the performance of Algorithm 1 and Algorithm 2 when nonlinear function approximators, for example, neural networks, are adopted. Although it seems difficult to establish convergence guarantees in this case, we believe that the empirical results are of independent interest, which justify the effectiveness of the proposed fully decentralized algorithms in a more sophisticated environment.

To this end, we consider the simulation environment of the *Cooperative Navigation* task in Lowe et al. (2017). In this environment, agents need to reach a set of L landmarks through physical movement. Agents are able to observe the position of the landmarks and other agents, and are rewarded based on the proximity of any agent to each landmark (Lowe et al., 2017). To be compatible with our networked multi-agent MDP, we modify the environment there in the following aspects. First, we assume the state is globally observable, i.e., the position of the landmarks and other agents are observable to each agent. Moreover, each agent has a certain target landmark to cover, and the individual reward is determined by the proximity to that certain landmark, as well as the penalty from collision with other agents. In this way, the reward function varies between agents. The reward is further scaled by different positive coefficients, representing the different priority/preferences of different agents. In addition, agents are connected via a time-varying communication network with several other agents nearby. The collaborative goal of the agents is then to maximize the network-wide averaged long-term return. The illustration of the modified Cooperative Navigation environment is provided in Figure 4.

Specifically, we consider $N = 10$ agents moving in a rectangular region of size $2 \times 2$. Each agent has a single target landmark, i.e., L = N = 10, which is randomly located in the region. The action set for each agent is the movement set {*left*, *right*, *up*, *down*, *stay*}, and thus $|\mathcal{A}_i| = 5$ for any $i \in \mathcal{N}$. The state $s$ includes the position of the landmarks and other agents, which thus has a dimension of $2(N + L) = 40$. The reward of agent $i$ is the negative number of the distance to the target landmark, plus $-1$ if agent $i$ collides with any other agents. The coefficients that scale the reward of each agent are selected randomly from a uniform distribution over $[0, 2]$. Each agent maintains two neural networks for actor and critic, respectively. Both neural networks have one hidden layer containing 24 neural units, which all use ReLU as the activation function. The output layer for the actor network is softmax, and that for the critic network is linear.

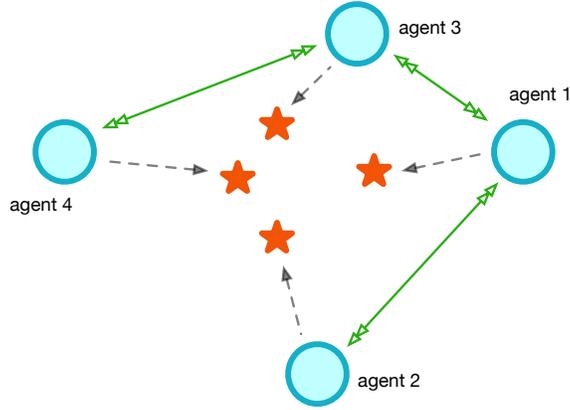

Figure 4: Illustration of the experimental environment for the Cooperative Navigation task we consider, modified from Lowe et al. (2017). In particular, the blue circles represent the agents, the orange stars represent the landmarks, the green arrows represent the communication links between agents, and the gray arrows show the target landmark each agent need to cover.

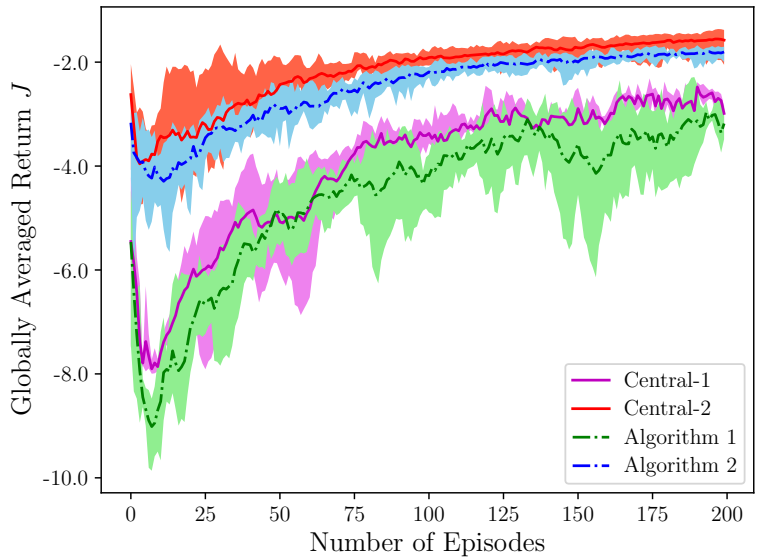

Figure 5: The globally averaged returns for the task of Cooperative Navigation, when neural networks are used for function approximation. We plot the returns achieved by both Algorithm 1 and Algorithm 2, along with their centralized counterparts Central-1 and Central-2.

The time-varying network $\mathcal{G}_t$ and consensus matrix $C_t$ are constructed in the same way as in §6.2. The stepsizes for the actor and critic step are set as constants 0.001 and 0.01, respectively. For each episode, the algorithms terminate either all agents reach the target landmarks or after 1000 iterations, and we run in total 200 episodes in each test run. We report the globally averaged return from 10 test runs in Figure 5.

It is shown in Figure 5 that the proposed algorithms successfully converge even with such nonlinear function approximators. Both decentralized algorithms are able to achieve globally averaged return close to the centralized counterparts, though at a slightly slower speed. This may also be explained by the delay of information diffusion across the network. We note that the AC algorithms based on state-value TD error, namely, Algorithm 2 and Central-2, seem to be superior to those based on action-value function approximation, namely, Algorithm 1 and Central-1. The former may achieve higher globally averaged reward than the latter with smaller variance. Our decentralized algorithms seem to inherit such convergence properties of their centralized counterparts. These observations justify the potential applicability of our algorithms to large-scale MARL problems, when neural networks are used as function approximators.

# 7 Conclusions

In this paper, we address the problem of multi-agent reinforcement learning with networked agents. In particular, we consider the *fully decentralized* setting where each agent makes individual decisions and receives local rewards, while exchanging information with neighbors over the network to accomplish optimal network-wide averaged return. Within this setting, we propose two decentralized actor-critic algorithms with function approximation, which can handle large-scale MARL problems with numerous agents and massive state-action spaces. We provide theoretical analysis on the convergence of the proposed algorithms with linear function approximation. An interesting direction of future research is to extend our algorithms and analyses to the setting with not only collaborative agents, but also competitive ones over the network. Moreover, it is also promising to extend the proposed algorithms to the MARL setting with continuous action spaces.

# A  Proof for the Stability of Consensus Updates

As mentioned before, the stability of the updates in stochastic approximation is usually proved separately. Here we provide the proof for the stability of the slower time scale update $\{\omega_t^i\}$ in Algorithm 1 and $\{z_t^i\}$ in Algorithm 2, i.e., Lemmas 5.1 and 5.4. In particular, we provide a sufficient condition for the stability of the consensus-based SA updates following the spirit of the results in Borkar and Meyn (2000) and Mathkar and Borkar (2016). We first state a main theorem for stability and then verify that Lemmas 5.1 and 5.4 are two special cases of it.

Let $\mathcal{N}$ be the set of agents with $|\mathcal{N}| = N$ and $x^i \in \mathbb{R}^d$ for any $i \in \mathcal{N}$. Consider the consensus update for $x_n^i \in \mathbb{R}^d$ as[6]

$$x_{n+1}^i = \sum_{j \in \mathcal{N}} c_n(i,j)\big\{x_n^j + \gamma_n\big[h^j(x_n, Y_n) + M_{n+1}^j\big]\big\}, \quad \text{for any } i \in \mathcal{N}, \tag{A.1}$$

where $\{Y_n\}_{n \geq 0}$ is an irreducible and aperiodic Markov chain over the finite set $A$. Let $\eta$ denote the stationary distribution of $\{Y_n\}$ and $\overline{h}^i(x_n) = \mathbb{E}_{Y_n \sim \eta}[h^i(x_n, Y_n)]$ denote the expectation of $h^i(x_n, Y_n)$ over $\eta$. Let $x_n = [(x_n^1)^\top, \cdots, (x_n^N)^\top]^\top \in \mathbb{R}^{dN}$, $C_n = [c_n(i,j)]_{N \times N}$, $h = [(h^1)^\top, \cdots, (h^N)^\top]^\top \in \mathbb{R}^{dN}$, $\overline{h}(x) = [(\overline{h}^1)^\top, \cdots, (\overline{h}^N)^\top]^\top \in \mathbb{R}^{dN}$, and $M_n = [(M_n^1)^\top, \cdots, (M_n^N)^\top]^\top \in \mathbb{R}^{dN}$. Let $\{\mathcal{F}_n\}$ be the filtration with $\mathcal{F}_n = \sigma(x_m, M_m, Y_m, C_{m-1}, m \leq n)$.

**Assumption A.1.** We make the following assumptions:

(a.1) The consensus weight matrices $\{C_n\}$ satisfy Assumption 4.3;

(a.2) $h^i : \mathbb{R}^n \times A \to \mathbb{R}^n$ is Lipschitz continuous in its first argument for any $i \in \mathcal{N}$;

(a.3) $\{M_n\}$ is a martingale difference sequence satisfying

$$\mathbb{E}\big(\|M_{n+1}\|^2 \,|\, \mathcal{F}_n\big) \leq K \cdot \big(1 + \|x_n\|^2\big),$$

for some $K > 0$;

(a.4) The difference $\zeta_{n+1} = \overline{h}(x_n) - h(x_n, Y_n)$ satisfies

$$\|\zeta_{n+1}\|^2 \leq K' \cdot \big(1 + \|x_n\|^2\big) \quad \text{a.s.,}$$

for some $K' > 0$;

(a.5) The stepsize sequence $\{\gamma_n\}$ satisfies $\sum_n \gamma_n = \infty$ and $\sum_n \gamma_n^2 < \infty$;

(a.6) Define $h_c : \mathbb{R}^{dN} \to \mathbb{R}^{dN}$ as $h_c(x) = \overline{h}(cx) \cdot c^{-1}$ with some $c > 0$, and $\widetilde{h}_c(y) : \mathbb{R}^d \to \mathbb{R}^d$ be $\widetilde{h}_c(y) = \langle h_c(\mathbb{1} \otimes y) \rangle$. Then $\widetilde{h}_c(y) \to h_\infty(y)$ as $c \to \infty$ uniformly on compact sets for some $h_\infty(y) : \mathbb{R}^d \to \mathbb{R}^d$. Also, for some $\epsilon < N^{-1/2}$, $B^\epsilon = \{y \,|\, \|y\| < \epsilon\}$ contains a globally asymptotically stable attractor of the ODE $\dot{y} = h_\infty(y)$.

---

[6]To avoid possible confusion with the notation of continuous time $t$ needed in the stability analysis, we use subscript $n$ to denote the iteration index in the proof of Theorem A.2.

Note that the definitions of $h_c$ and $h_\infty$ in (a.6) in Assumption A.1 are different from those in Mathkar and Borkar (2016). Here we consider the averaged ODE in the consensus subspace for each agent, while that reference considers the overall ODE associated with (A.1), i.e., define $\widetilde{h} : \mathbb{R}^{dN} \to \mathbb{R}^{dN}$ as $\widetilde{h}(x) = C_* \bar{h}(x)$ and let $h_\infty(x) = \lim_c \widetilde{h}(cx) \cdot c^{-1}$, where $C_* = \lim_n \prod_{m=1}^n C_m$. In fact, from Nedic and Ozdaglar (2009); Nedich et al. (2016), the limit $C_*$ exists and has identical rows and rank one, provided the sequence $\{C_t\}$ satisfies Assumption 4.3. Therefore, the globally asymptotical stability of the ODE $\dot{x} = h_\infty(x)$ (see Assumption (A5) in Mathkar and Borkar (2016)) does not hold for the linear ODE we consider in the convergence proof of the critic steps in both algorithms. In contrast, we can verify our condition (a.6) in Assumption A.1 later in the proof of Lemmas 5.1 and 5.4. We then have the following theorem on the stability of the sequence $\{x_n\}$.

**Theorem A.2.** *Under Assumption A.1, the sequence $\{x_n\}$ generated from (A.1) is bounded almost surely, i.e., $\sup_n \|x_n^i\| < \infty$ a.s. for any $i \in \mathcal{N}$.*

## A.1 Proof of Theorem A.2

Let $\vartheta_c(y, t)$ denote the solution to the ODE

$$\dot{y} = \widetilde{h}_c(y), \; y(0) = y \tag{A.2}$$

We first have the following lemma, which is similar to Lemma 5 in Mathkar and Borkar (2016), and thus we do not provide its proof here.

**Lemma A.3.** *There exist constants $c_0 > 0$ and $T > 0$ such that for all initial conditions $y$ within the sphere $\{y \mid \|y\| \le N^{-1/2}\}$ and all $c \ge c_0$, we have $\|\vartheta_c(y, t)\| \le (1 - \epsilon') \cdot N^{-1/2}$ for $t \in [T, T+1]$, for some $0 < \epsilon' < 1$, where $N$ is the number of agents.*

Now, before stating the next set of lemmas, we introduce some notations and terminology. First, by the convention adopted in Borkar and Meyn (2000), we define $t_0 = 0$ and $t_n = \sum_{i=0}^n \gamma_i, n \ge 0$. Then we define $\bar{x}(t), t \ge 0$ as $\bar{x}(t_n) = x_n, n \ge 0$ with linear interpolation on each interval $[t_n, t_{n+1}]$. Moreover, we let $T_0 = 0$ and $T_{n+1} = \min\{t_m : t_m \ge T_n + T\}$ for any $n \ge 0$. Then $T_{n+1} \in [T_n + T, T_n + T + \sup_n \gamma_n]$. Let $m(n)$ be such that $T_n = t_{m(n)}$, for $n \ge 0$. Define the piecewise continuous trajectory $\hat{x}(t) = \bar{x}(t) \cdot r_n^{-1}$ for $t \in [T_n, T_{n+1})$, where $r_n = \max_{i \in \mathcal{N}} \{\|\bar{x}(T_n)\|, 1\}$. This implies that $\|\hat{x}(T_n)\| \le 1$ for any $n \ge 0$. We also define $\hat{x}(T_{n+1}^-) = \bar{x}(T_{n+1}) \cdot r_n^{-1}$, $\hat{M}_{k+1} = M_{k+1} \cdot r_n^{-1}$, and $\hat{\zeta}_{k+1} = \zeta_{k+1} \cdot r_n^{-1}$ for $k \in [m(n), m(n+1))$.

Note that $\{\hat{M}_k\}$ is also a martingale difference sequence as $\{M_k\}$. We first establish boundedness of $\mathbb{E}[\|\hat{x}(t)\|^2]$ as follows.

**Lemma A.4.** *Under Assumption A.1, $\sup_t \mathbb{E}[\|\hat{x}(t)\|^2] < \infty$.*

*Proof.* It suffices to show that $\sup_{m(n) \le k < m(n+1)} \mathbb{E}[\|\hat{x}(t_k)\|^2] < M$ for some $M > 0$ independent of $n$. We first write the update of $\hat{x}(t_k)$ for $k \in [m(n), m(n+1))$ in a compact form as

$$\hat{x}(t_{k+1}) = (C_k \otimes I)\Big(\hat{x}(t_k) + \gamma_k \big\{h_{r_n}[\hat{x}(t_k)] + \hat{M}_{k+1} + \hat{\zeta}_{k+1}\big\}\Big). \tag{A.3}$$

Note that the additional term $\hat{\zeta}_{k+1}$ also satisfies $\mathbb{E}(\|\hat{\zeta}_{k+1}\|^2 | \mathcal{F}_k) \le K' \cdot (1 + \|\hat{x}(t_k)\|^2)$ by condition (a.4) in Assumption A.1 since $r_n \ge 1$. Moreover, since $C_k \otimes I$ has bounded norm, it follows similarly as in the proof for Lemma 4 on page 25 in Borkar (2008) that

$$\mathbb{E}[\|\hat{x}(t_{k+1})\|^2]^{1/2} \le \mathbb{E}[\|\hat{x}(t_k)\|^2]^{1/2}(1 + \gamma_k K_1) + \gamma_k K_2,$$

for some $K_1, K_2 > 0$ and $k \in [m(n), m(n+1))$. Then, by Grönwall inequality, we have the desired boundedness of $\mathbb{E}[\|\hat{x}(t)\|^2]$. $\square$

By Lemma A.4, we immediately have the following result.

**Lemma A.5** (Lemma 5 on page 25 in Borkar (2008)). *The sequence $\{\sum_{k=0}^{n-1} \gamma_k \hat{M}_{k+1}\}$ converges almost surely.*

We thus obtain the almost sure boundedness of the trajectory $\{\hat{x}(t)\}$.

**Lemma A.6.** *Under Assumption A.1, $\sup_t \|\hat{x}(t)\| < \infty$ a.s.*

*Proof.* Recall the update in (A.3) and note that $\|(C_k \otimes I)x\|_\infty \leq \|x\|_\infty$ since $C_k$ is a row stochastic matrix. Thus we have

$$\|\hat{x}(t_{k+1})\|_\infty \leq \|\hat{x}(t_k)\|_\infty + \gamma_k \left\| h_{r_n}[\hat{x}(t_k)] + \hat{M}_{k+1} + \hat{\zeta}_{k+1} \right\|_\infty. \tag{A.4}$$

By iterating (A.4), we obtain

$$\|\hat{x}(t_{k+1})\|_\infty \leq \|\hat{x}(t_{m(n)})\|_\infty + \sum_{l=0}^{k-m(n)} \gamma_{m(n)+l} \left( \left\| h_{r_n}[\hat{x}(t_{m(n)+l})] \right\|_\infty + \left\| \hat{M}_{m(n)+l+1} \right\|_\infty + \left\| \hat{\zeta}_{m(n)+l+1} \right\|_\infty \right)$$

$$\leq \|\hat{x}(t_{m(n)})\|_\infty + \sum_{l=0}^{k-m(n)} \gamma_{m(n)+l} \cdot K_3 \left[ 1 + \left\| \hat{x}(t_{m(n)+l}) \right\|_\infty \right] + \sum_{l=0}^{k-m(n)} \gamma_{m(n)+l} \left\| \hat{M}_{m(n)+l+1} \right\|_\infty, \tag{A.5}$$

for some $K_3 > 0$. The second inequality is due to the Lipschitz continuity of $h_{r_n}$ and condition (a.4) on $\zeta_{n+1}$, by the equivalence of vector norms. Moreover, by Lemma A.5, the third term on the RHS of (A.5) is bounded a.s. since $\{\sum_{k=0}^{n-1} \gamma_k \hat{M}_{k+1}\}$ converges a.s. Recall that $\sum_{l=0}^{k-m(n)} \gamma_{m(n)+l} \leq T + \sup_n \gamma_n < \infty$ by definition of $m(n)$ and $T_n$. Thus, there exist $K_4, K_5 > 0$ such that

$$\|\hat{x}(t_{k+1})\|_\infty \leq K_4 + K_5 \sum_{l=0}^{k-m(n)} \gamma_{m(n)+l} \left\| \hat{x}(t_{m(n)+l}) \right\|_\infty.$$

By discrete-time Grönwall inequality, we have

$$\sup_{m(n) \leq k < m(n+1)} \|\hat{x}(t_k)\|_\infty \leq K_4 \cdot \exp\left[ K_5 \cdot \left( T + \sup_n \gamma_n \right) \right], \tag{A.6}$$

where the RHS of (A.6) is a (random) constant independent of $n$. Hence by equivalence of vector norms, we further obtain $\sup_t \|\hat{x}(t)\| < \infty$, which concludes the proof. $\square$

The stability of $\|\hat{x}(t)\|$ is essential in showing the convergence of the consensus update in (A.3). For $n \geq 0$, let $y^n(t)$ denote the trajectory of $\dot{y} = \widetilde{h}_c(y)$ with $c = r_n$ and $y^n(T_n) = \langle \hat{x}(T_n) \rangle$, for $t \in [T_n, T_{n+1})$. Then we have the following lemma.

**Lemma A.7.** *Under Assumption A.1, $\lim_n \sup_{t \in [T_n, T_{n+1})} \|\hat{x}(t) - \mathbb{1} \otimes y^n(t)\| = 0$.*

*Proof.* Since $\hat{x}(t)$ is bounded a.s. on $[T_n, T_{n+1})$, we can mimic our proofs for Theorems 4.6 and 4.10 to show the convergence of $\hat{x}(t_k)$ for $k \in [T_n, T_{n+1}]$ as $n \to \infty$. We will provide here only a sketch. One can first show that over the set $\{\sup_k \|\hat{x}(t_k)\| \leq M\}$ for any $M > 0$, $\lim_k \|\mathcal{J}_\perp \hat{x}(t_k)\| = 0$. The iteration of $\mathcal{J}_\perp \hat{x}(t_k)$ has the form (similar to (5.6))

$$\mathcal{J}_\perp \hat{x}(t_{k+1}) = [(I - \mathbb{1}\mathbb{1}^\top/N)C_k \otimes I][\mathcal{J}_\perp \hat{x}(t_k) + \gamma_k y_{k+1}], \tag{A.7}$$

where $y_{k+1} = h_{r_n}[\hat{x}(t_k)] + \hat{M}_{k+1} + \hat{C}_{k+1}$ here. One can easily verify that $\mathbb{E}(\|y_{k+1}\|^2 \mathbb{I}_{\{\sup_k \|\hat{x}(t_k)\| \leq M\}} | \mathcal{F}_k) < K_6$ for some $K_6 > 0$, due to Lipschitz continuity of $h_{r_n}$ and the conditions (a.3) and (a.4) in Assumption A.1. Hence, by similar arguments as in the proof of Lemma 5.3, we obtain $\lim_k \|\mathcal{J}_\perp \hat{x}(t_k)\| = 0$ almost surely, i.e., the vector $\hat{x}(t_k)$ reaches consensus as $k \to \infty$. Then we proceed to show the convergence of the sequence $\{\langle \hat{x}(t_k) \rangle\}$. Define $\hat{h}_c : \mathbb{R}^d \times A \to \mathbb{R}^d$ as $\hat{h}_c(y, Y_k) = \langle h(c \cdot \mathbb{1} \otimes y) \cdot c^{-1} \rangle$; then the iteration can be written as follows

$$\langle \hat{x}(t_{k+1}) \rangle = \langle \hat{x}(t_k) \rangle + \gamma_k \cdot \mathbb{E}(\langle y_{k+1} \rangle | \mathcal{F}_k) + \gamma_k \cdot \xi_{k+1} = \langle \hat{x}(t_k) \rangle + \gamma_k \cdot \hat{h}_{r_n}[\langle \hat{x}(t_k) \rangle, Y_k] + \gamma_k \cdot \xi_{k+1} + \gamma_k \cdot \beta_{k+1},$$

where $\xi_{k+1} = \langle (C_k \otimes I)(y_{k+1} + \gamma_k^{-1} \mathcal{J}_\perp \hat{x}(t_k)) \rangle - \mathbb{E}(\langle y_{k+1} \rangle | \mathcal{F}_k)$, $\beta_{k+1} = \mathbb{E}(\langle y_{k+1} \rangle | \mathcal{F}_k) - \hat{h}_{r_n}[\langle \hat{x}(t_k) \rangle, Y_k]$, and $\langle y_{k+1} \rangle = \langle h_{r_n}[\hat{x}(t_k)] \rangle + \hat{M}_{k+1} + \langle \hat{C}_{k+1} \rangle$. One can verify that $\{\xi_{k+1}\}$ is a martingale difference sequence satisfying $\mathbb{E}(\|\xi_{t+1}\|^2 | \mathcal{F}_{t,1}) < K_7 \cdot (1 + \|\langle \hat{x}(t_k) \rangle\|^2)$ for some $K_7 < \infty$ over the set $\{\sup_k \|\hat{x}(t_k)\| \leq M\}$. In addition, note that $\mathbb{E}(\hat{M}_{k+1} | \mathcal{F}_k) = 0$ and thus $\mathbb{E}(\langle y_{k+1} \rangle | \mathcal{F}_k) = \langle h[\hat{x}(t_k), Y_k] \rangle \cdot r_n^{-1}$. Thus we have $\|\beta_{k+1}\| \leq L \cdot \|\mathcal{J}_\perp \hat{x}(t_k)\| \cdot r_n^{-1}$ for some $L < \infty$ due to the Lipschitz continuity of $h$. Hence $\beta_k \to 0$ a.s. since $\|\mathcal{J}_\perp \hat{x}(t_k)\| \to 0$ a.s. and $r_n \geq 1$. Moreover, $\hat{h}_{r_n}[\langle \hat{x}(t_k) \rangle, Y_k]$ is Lipschitz continuous in $\langle \hat{x}(t_k) \rangle$. Therefore, by Theorem B.2, we obtain that $\langle \hat{x}(t_k) \rangle \to y^n(t)$ as $n \to \infty$, namely $k \to \infty$. Further we obtain that $\hat{x}^i(t_k) \to y^n(t)$ for any $i \in \mathcal{N}$, which concludes the proof following Theorem 2 in Chapter 2 of Borkar (2008). □

Now suppose that $\|x_n^i\| \to \infty$ for some $i \in \mathcal{N}$; then there exists a subsequence of $\{n_q\}$ such that $\|\overline{x}^i(T_{n_q})\| \to \infty$ and thus $\|\overline{x}(T_{n_q})\| \to \infty$. Hence $r_{n_q} \to \infty$. If $r_n > c_0 \geq 1$, then $\|\hat{x}(T_n)\| = 1$, and thus $\|y^n(T_n)\| = \|\langle \hat{x}(T_n) \rangle\| \leq N^{-1/2}$. By Lemma A.3, we have $\|\mathbb{1} \otimes y^n(T_n^-)\| = N^{1/2} \cdot \|y^n(T_n^-)\| \leq 1 - \epsilon'$. Thus by Lemma A.7, we have $\|\hat{x}(T_{n+1}^-)\| < 1 - \epsilon''$ for some $0 < \epsilon'' < \epsilon'$. Hence for $r_n > c_0$ and sufficiently large $n$,

$$\frac{\|\overline{x}(T_{n+1})\|}{\|\overline{x}(T_n)\|} = \frac{\|\hat{x}(T_{n+1}^-)\|}{\|\hat{x}(T_n)\|} < 1 - \epsilon''.$$

It thus follows that if $\|\overline{x}(T_n)\| > 1$, $\|\overline{x}(T_k)\|$ falls back to the unit ball at an exponential rate for $k \geq n$. The rest of the argument follows directly from the proof of Theorem 2 in Mathkar and Borkar (2016), which concludes the proof. □

Now we are ready to apply Theorem A.2 to prove Lemmas 5.1 and 5.4. We will return to the notations in §4.

## A.2 Proof of Lemma 5.1

The proof follows by verifying the conditions for Theorem A.2 to hold. Recall that the critic step in (3.7) has the form

$$\omega_{t+1} = (C_t \otimes I)(\omega_t + \beta_{\omega,t} \cdot y_{t+1}),$$

with $y_{t+1} = (\delta_t^1 \phi_t^\top, \cdots, \delta_t^N \phi_t^\top)^\top \in \mathbb{R}^{KN}$. Thus the terms corresponding to (A.1) are

$$h^i(\omega_t^i, \mu_t^i, s_t, a_t) = \mathbb{E}(\delta_t^i \phi_t^\top | \mathcal{F}_{t,1}), \quad M_{t+1}^i = \delta_t^i \phi_t^\top - \mathbb{E}(\delta_t^i \phi_t^\top | \mathcal{F}_{t,1}). \tag{A.8}$$

Since the Markov chain $\{(s_t, a_t)\}_{t \geq 0}$ is irreducible and aperiodic given policy $\pi_\theta$, we have $\overline{h}^i(\omega_t^i, \mu_t^i) = \Phi^\top D_\theta^{s,a}[R^i - \mu_t^i \mathbb{1} + P^\theta \Phi \omega_t^i - \Phi \omega_t^i]$. By Lemma 5.2, it is established that $\{\mu_t^i\}$ is bounded a.s. Hence, over the set $\{\sup_t \|\mu_t\| \leq M\}$ for any $M > 0$, there exists some $K' > 0$ such that $\|\bar{h}(\omega_t, \mu_t) - h(\omega_t, \mu_t, s_t, a_t)\|^2 \leq K' \cdot (1 + \|\omega_t\|^2)$, since the Markov chain is finite. This verifies the condition (a.4) in Assumption A.1. Moreover, since $r^i_{t+1}$ and $\|\phi_t\|$ are uniformly bounded, $\mathbb{E}(\|M_{t+1}\|^2 | \mathcal{F}_{t,1}) \leq K \cdot (1 + \|\omega_t\|^2)$ is also verified for some $K > 0$. More importantly, over the set $\{\sup_t \|\mu_t\| \leq M\}$, $h_\infty(y)$ exists and has the form

$$h_\infty(y) = \lim_c \widetilde{h}_c(y) = \Phi^\top D_\theta^{s,a}(P^\theta - I)\Phi y. \tag{A.9}$$

Clearly $\dot{y} = h_\infty(y)$ has origin as its globally asymptotically stable attractor (see the proof of Theorem 4.6 in §5.1). Hence we apply Theorem A.2 to conclude the proof. $\square$

## A.3 Proof of Lemma 5.4

Recall that the critic step from (3.12), (3.13), and (3.16) has the compact form

$$z_{t+1} = (C_t \otimes I)\bigl(z_t + \beta_{v,t} \cdot y_{t+1}\bigr), \tag{A.10}$$

where $z^i_t = [\mu^i_t, (\lambda^i_t)^\top, (v^i_t)^\top]^\top$ and $y_t = [(y^1_t)^\top, \cdots, (y^N_t)^\top]^\top \in \mathbb{R}^{(1+M+L)N}$. Here $y^i_{t+1}$ denotes $y^i_{t+1} = [r^i_{t+1} - \mu^i_t, (r^i_{t+1} - f_t^\top \lambda^i_t) f_t^\top, \delta^i_t \varphi_t^\top]^\top$. Thus the terms corresponding to (A.1) are

$$h^i(z^i_t, s_t, a_t) = \mathbb{E}(y^i_{t+1} | \mathcal{F}_{t,1}), \quad M^i_{t+1} = y^i_{t+1} - \mathbb{E}(y^i_{t+1} | \mathcal{F}_{t,1}). \tag{A.11}$$

Furthermore, we have

$$\bar{h}^i(z^i_t) = \begin{pmatrix} -1 & 0 & 0 \\ 0 & -F^\top D_\theta^{s,a} F & 0 \\ -\Phi^\top D_\theta^s \mathbb{1} & 0 & \Phi^\top D_\theta^s(P^\theta - I)\Phi \end{pmatrix} \begin{pmatrix} \mu^i_t \\ \lambda^i_t \\ v^i_t \end{pmatrix} + \begin{pmatrix} J^i(\theta) \\ F^\top D_\theta^{s,a} R^i \\ \Phi^\top D_\theta^s R^i_\theta \end{pmatrix},$$

where $J^i(\theta) = \sum_{s \in \mathcal{S}, a \in \mathcal{A}} d_\theta(s,a) \cdot R^i(s,a)$ and $R^i_\theta(s) = \sum_a \pi_\theta(s,a) R^i(s,a)$. Therefore, one can verify that both conditions (a.3) and (a.4) in Assumption A.1 are satisfied. In addition, $h_\infty(y)$ exists and has the form

$$h_\infty(y) = \lim_c \widetilde{h}_c(y) = \begin{pmatrix} -1 & 0 & 0 \\ 0 & -F^\top D_\theta^{s,a} F & 0 \\ -\Phi^\top D_\theta^s \mathbb{1} & 0 & \Phi^\top D_\theta^s(P^\theta - I)\Phi \end{pmatrix} \cdot y.$$

Clearly $\dot{y} = h_\infty(y)$ has origin as its globally asymptotically stable attractor (see the proof of Theorem 4.10 in §5.3), which completes the proof. $\square$

# B Technical Background

## B.1 A Basic Result of Stochastic Approximation

For the sake of completeness, we reproduce here a key result from Borkar (2008) that has been used repeatedly in our proofs. The results follow by specializing Corollary 8 and Theorem 9 on page 74-75 in Borkar (2008). We note that this is actually an extension of Theorem 2.1 and Theorem 2.2 in Borkar and Meyn (2000) to the case with irreducible Markovian state and diminishing noise in the update. More general conclusions can also be found in Benaïm (1999); Kushner and Yin (2003).

Consider the $n$-dimensional stochastic approximation iteration

$$x_{t+1} = x_t + \gamma_t [h(x_t, Y_t) + M_{t+1} + \beta_{t+1}], \ t \geq 0, \tag{B.1}$$

where $\gamma_t > 0$ and $\{Y_t\}_{t \geq 0}$ is a Markov chain on a finite set $A$.

**Assumption B.1.** We make the following assumptions:

(a.1) $h : \mathbb{R}^n \times A \to \mathbb{R}^n$ is Lipschitz continuous in its first argument;

(a.2) $\{Y_t\}_{t \geq 0}$ is an irreducible Markov chain with stationary distribution $\pi$;

(a.3) The stepsize sequence $\{\gamma_t\}$ satisfies $\sum_t \gamma_t = \infty$ and $\sum_t \gamma_t^2 < \infty$;

(a.4) $\{M_t\}$ is a martingale difference sequence, i.e., $\mathbb{E}[M_{t+1} | x_\tau, M_\tau, Y_\tau, \tau \leq t] = 0$, satisfying that for some $K > 0$ and $t \geq 0$

$$\mathbb{E}\big(\|M_{t+1}\|^2 \,\big|\, x_\tau, M_\tau, Y_\tau, \tau \leq t\big) \leq K \cdot \big(1 + \|x_t\|^2\big).$$

(a.5) The sequence $\{\beta_t\}$ is a bounded random sequence with $\beta_t \to 0$ almost surely as $t \to \infty$.

Then the asymptotic behavior of the iteration (B.1) is related to the behavior of the solution to the ODE

$$\dot{x} = \overline{h}(x) = \sum_i \pi(i) h(x, i). \tag{B.2}$$

Suppose (B.2) has a unique globally asymptotically stable equilibrium $x^*$, we then have the following two theorems.

**Theorem B.2.** Under Assumption B.1, if $\sup_t \|x_t\| < \infty$ a.s., we have $x_t \to x^*$.

**Theorem B.3.** Under Assumption B.1, suppose that

$$\lim_{c \to \infty} \frac{\overline{h}(cx)}{c} = h_\infty(x)$$

exists uniformly on compact sets for some $h_\infty \in C(\mathbb{R}^n)$. If the ODE $\dot{y} = h_\infty(y)$ has origin as the unique globally asymptotically stable equilibrium, then

$$\sup_t \|x_t\| < \infty \ \text{ a.s.}$$

## B.2 Kushner-Clark Lemma

We state here the well-known Kushner-Clark Lemma (Kushner and Clark, 1978; Metivier and Priouret, 1984; Prasad et al., 2014) in the sequel.

Let $\Gamma$ be an operator that projects a vector onto a compact set $\mathcal{X} \subseteq \mathbb{R}^N$. Define a vector $\hat{\Gamma}(\cdot)$ as

$$\hat{\Gamma}[h(x)] = \lim_{0 < \eta \to 0} \left\{ \frac{\Gamma[x + \eta h(x)] - x}{\eta} \right\},$$

for any $x \in \mathcal{X}$ and with $h : \mathcal{X} \to \mathbb{R}^N$ continuous. Consider the following recursion in $N$ dimensions

$$x_{t+1} = \Gamma\{x_t + \gamma_t [h(x_t) + \xi_t + \beta_t]\}. \tag{B.3}$$

The ODE associated with (B.3) is given by

$$\dot{x} = \hat{\Gamma}[h(x)]. \tag{B.4}$$

**Assumption B.4.** We make the following assumptions:

(a.1) $h(\cdot)$ is a continuous $\mathbb{R}^N$-valued function.

(a.2) The sequence $\{\beta_t\}$, $t \geq 0$ is a bounded random sequence with $\beta_t \to 0$ almost surely as $t \to \infty$.

(a.3) The stepsizes $\gamma_t$, $t \geq 0$ satisfy $\gamma_t \to 0$ as $t \to \infty$ and $\sum_t \gamma_t = \infty$.

(a.4) The sequence $\xi_t$, $t \geq 0$ satisfies for any $\epsilon > 0$

$$\lim_t \mathbb{P}\left( \sup_{n \geq t} \left\| \sum_{\tau=t}^n \gamma_\tau \xi_\tau \right\| \geq \epsilon \right) = 0.$$

Then the Kushner-Clark Lemma says the following.

**Theorem B.5.** *Under Assumption B.4, suppose that the ODE (B.4) has a compact set $\mathcal{K}^*$ as its set of asymptotically stable equilibria. Then $x_t$ in (B.3) converges almost surely to $\mathcal{K}^*$ as $t \to \infty$.*

# C Comparison with Existing Work on Multi-Agent Systems and MARL

In this section, we compare both our *model* and *algorithms* with related work on multi-agent systems and collaborative MARL in details.

Our framework of networked multi-agent systems finds a broad range of applications in distributed cooperative control problems, including formation control of unmanned vehicles (Fax and Murray, 2004), cooperative navigation of robots (Corke et al., 2005), load management in energy networks (Dall'Anese et al., 2013), and flocking of mobile sensor networks (Cortes et al., 2004), etc. Previously, the collective goal of the multi-agent system is to either reach a stable and consensus state for all agents (Fax and Murray, 2004; Corke et al., 2005), or solve a static optimization problem in a distributed fashion (Dall'Anese et al., 2013; Nedic and Ozdaglar, 2009). In the first line of work, including formation control and consensus problems, the objective is not formulated explicitly as an optimization problem, and most of the work focuses on continuous-time dynamic systems. Whereas in the second line of work, the problem is approached in a static setting, in the sense that the optimization objective is deterministic and there is no control input affecting the transition of the system, see recent efforts in Nedic and Ozdaglar (2009); Agarwal and Duchi (2011); Jakovetic et al. (2011); Tu and Sayed (2012); Chen and Sayed (2012). In contrast, we here model the interaction of multiple agents and evolution of the system as an MDP, a dynamic setting, and explicitly use the network-wide long-term return as the collaborative goal of all agents. In this regard, our framework is pertinent to the cooperative/distributed optimal control problems (Lewis et al., 2013; Movric and Lewis, 2014), but focuses on the discrete-time setting and falls into the realm of reinforcement learning, where the model of the system may be unknown. One recent work (Macua et al., 2017) for multi-task RL, which is almost concurrent to ours, is also based on the model with networked agents. Nonetheless, the MDP problem solved by different agents are totally decoupled, which excludes the work from the realm of MARL with interactive agents as we consider here.

Our framework also departs from the existing framework on collaborative MARL models in the following aspects. In contrast to the canonical multi-agent MDP (MMDP) model proposed in Boutilier (1996); Lauer and Riedmiller (2000), our model allows the agents to exchange information over a communication network with possibly sparse connectivity at each agent. This improves the *scalability* of the multi-agent model with a high population of agents, which is one of the long-standing challenges in general MARL problems (Shoham et al., 2003). Moreover, we allow heterogeneous agents to have various individual reward functions, while the canonical MMDP assumes a common reward function for all agents. The latter setting greatly simplifies the problem since no information exchange among agents is necessary to approximate the value function for each agent. Our model not only fits in the *multi-task* setting which has gained increasing popularity in MARL (Omidshafiei et al., 2017; Teh et al., 2017), but also applies to the general multi-agent RL setting. One of the few models that also consider heterogeneous reward functions in collaborative MARL is Kar et al. (2013), where the global action is assumed to be actuated by a remote controller, but in our case, the agents are fully decentralized and have local control capabilities. It is also worth noting that our model generalizes the team Markov game model for collaborative MARL, see Littman (2001); Wang and Sandholm (2003); Arslan and Yüksel (2017), where all agents have individual action sets but share a common payoff function as in the canonical MMDP.

Moreover, our algorithms designed for networked MMDP are distinct from the existing collaborative MARL algorithms in the following aspects. First, our MARL algorithms belong to the type of actor-critic algorithms, whereas several of the existing MARL algorithms are designed based on Q-learning type (critic-based) algorithms only (Boutilier, 1996; Lauer and Riedmiller, 2000; Kar et al., 2013). Moreover, these algorithms assume either the rewards are common to all agents (Boutilier, 1996; Lauer and Riedmiller, 2000), or there exists a remote central controller to take actions for the agents (Kar et al., 2013). More recently, some actor-critic type MARL algorithms with *distributed/decentralized* structures have gained increasing attention (Gupta et al., 2017; Lowe et al., 2017; Omidshafiei et al., 2017). They are developed for more complicated settings where both cooperation and competition may appear among agents. However, they all rely on a central controller to perform the critic step, which are closer to the *hierarchical* structure rather than the *fully decentralized* structure we consider here. Second, our algorithms apply function approximation to handle the setting with massively large state and action spaces, while enjoying theoretical guarantees for convergence as we show in §4. However, the existing collaborative MARL algorithms are either guaranteed to converge only for tabular cases (Hu and Wellman, 2003; Wang and Sandholm, 2003; Kar et al., 2013; Prasad et al., 2014), or only have empirical convergence when function approximation is applied (Foerster et al., 2016; Gupta et al., 2017; Lowe et al., 2017; Omidshafiei et al., 2017; Foerster et al., 2017; Lanctot et al., 2017). The recent work on multi-task RL with networked agents (Macua et al., 2017) also focuses on empirical results only, with no complete convergence analysis.